\documentclass[10pt,twocolumn,letterpaper]{article}
\usepackage[pagenumbers]{cvpr}
\definecolor{cvprblue}{rgb}{0.21,0.49,0.74}
\usepackage[pagebackref,breaklinks,colorlinks,allcolors=cvprblue]{hyperref}
\usepackage{graphicx}
\usepackage{subcaption}
\usepackage{tabularx}
\usepackage{array}
\usepackage{makecell}
\usepackage{multirow}
\usepackage{caption}
\usepackage{subcaption}
\usepackage[norelsize, linesnumbered, ruled, lined, boxed, commentsnumbered]{algorithm2e}

\title{Enhancing MMDiT-Based Text-to-Image Models for Similar Subject Generation}

\author{ Tianyi Wei\textsuperscript{\rm 1}, Dongdong Chen\textsuperscript{\rm 2}, Yifan Zhou\textsuperscript{\rm 1}, Xingang Pan\textsuperscript{\rm 1}\\
	\normalsize\textsuperscript{\rm 1}S-Lab, Nanyang Technological University  \ \normalsize\textsuperscript{\rm 2}Microsoft GenAI  \  \\
	{\tt\small\{tianyi.wei, yifan006, xingang.pan\}@ntu.edu.sg }, {\tt\small cddlyf@gmail.com} \\
}

\begin{document}

\twocolumn[{
	\renewcommand\twocolumn[1][]{#1}
	\maketitle
	\setlength\tabcolsep{0.5pt}
	\centering
	\small
	\begin{tabular}{c}
		\includegraphics[width=0.99\textwidth]{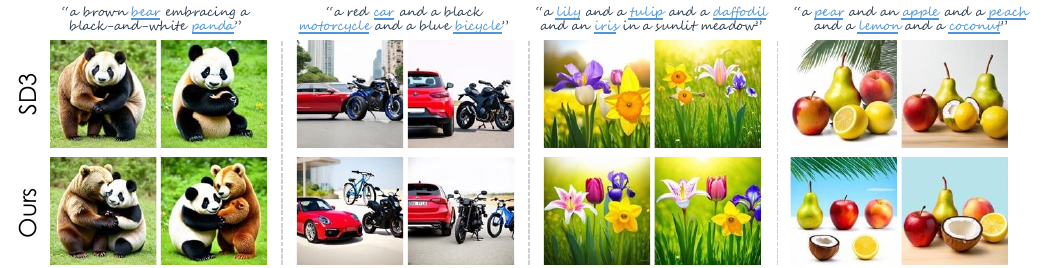}
	\end{tabular}
	\captionof{figure}{Our approach can effectively mitigate the subject neglect or mixing issues suffered by SD3 for similar subject generation.}
        \vspace{1.2em}
	\label{fig:teaser}
}]

\maketitle
\begin{abstract}
Representing the cutting-edge technique of text-to-image models, the latest Multimodal Diffusion Transformer (MMDiT) largely mitigates many generation issues existing in previous models. However, we discover that it still suffers from subject neglect or mixing when the input text prompt contains multiple subjects of similar semantics or appearance. We identify three possible ambiguities within the MMDiT architecture that cause this problem: Inter-block Ambiguity, Text Encoder Ambiguity, and Semantic Ambiguity. To address these issues, we propose to repair the ambiguous latent on-the-fly by test-time optimization at early denoising steps. In detail, we design three loss functions: Block Alignment Loss, Text Encoder Alignment Loss, and Overlap Loss, each tailored to mitigate these ambiguities. Despite significant improvements, we observe that semantic ambiguity persists when generating multiple similar subjects, as the guidance provided by overlap loss is not explicit enough. Therefore, we further propose Overlap Online Detection and Back-to-Start Sampling Strategy to alleviate the problem. Experimental results on a newly constructed challenging dataset of similar subjects validate the effectiveness of our approach, showing superior generation quality and much higher success rates over existing methods. Our code will be available at \url{https://github.com/wtybest/EnMMDiT}.
\end{abstract}
    
\section{Introduction}
\label{sec:intro}

In recent years, text-to-image generation models have advanced rapidly. Among them, the latent diffusion model series~\cite{rombach2022high}, as a leading open-source framework, has evolved from Stable Diffusion 1, 2, and SDXL~\cite{podell2023sdxl} to the latest Stable Diffusion 3~\cite{esser2024scaling}. Compared to its predecessors, Stable Diffusion 3 introduces several key improvements that further push the boundaries of text-to-image synthesis. These enhancements include a rectified flow formulation~\cite{liuflow}, a novel Multimodal Diffusion Transformer (MMDiT)~\cite{peebles2023scalable}  architecture, ensemble textual context information (encoded by two CLIP~\cite{radford2021learning} models and a T5~\cite{raffel2020exploring} model), and a new information interaction manner, where image and text tokens are concatenated and processed via self-attention.

With these substantial improvements, SD3 significantly mitigates generation issues encountered in previous versions, such as subject neglect and incorrect attribute binding~\cite{chefer2023attend}. However, we observe a significant limitation that remains: when the input text contains multiple subjects of similar semantics or appearance, SD3 still suffers from subject neglect or mixing as shown in Figure~\ref{fig:teaser}, leading to a significant decrease of its generation success rate (even less than $20\%$). To understand the underlying reasons for this issue, we visualize the cross-attention portion of the joint self-attention layer of the image text tokens, which provides reliable clues about the relationship between the generated pixels and different text tokens.

Interestingly, our investigation reveals that the MMDiT-based text-to-image model represented by Stable Diffusion 3 suffers from the following three types of ambiguities in the generation process, which cause the aforementioned generation issue: 1) \textit{Inter-block Ambiguity}. At each step in the denoising process, the recognition of the subject by the MMDiT blocks is progressively clearer. Although the blocks in later layers are more accurate in recognizing the subject, the effect of inaccurate attention activation in earlier layers has already been incurred, which leads to semantic leakage; 2) \textit{Text Encoder Ambiguity}. Taking ``a cat and a dog" as an example, the activated cross-attention of subject text representations (\ie, cat and dog) from the CLIP text encoder and T5 text encoder are sometimes inconsistent; 3) \textit{Semantic Ambiguity in Similar Subjects Themselves}. Some similar subjects such as ``a duck and a goose" are very similar in shape and appearance, which makes it difficult to distinguish them clearly at the early stage of denoising, thus resulting in the phenomenon that the attentions of these subjects coalesce in the same position.

Based on the above observations, we propose to repair the ambiguity on-the-fly by test-time optimization of the latent at the early denoising stage with the guidance from cross-attention. Accordingly, three kinds of losses are proposed: 1) \textit{Block Alignment Loss}. We propose to mitigate the inter-block ambiguity by self-refinement, \ie, the average subject attention information obtained by the later blocks is used to guide the previous layer's transform block to align with it, so as to attenuate the semantic leakage caused by the previous layer's blocks; 2) \textit{Text Encoder Alignment Loss}. Since it is difficult to determine which text encoder's activation is definitely correct, we employ implicit constraints to motivate the CLIP text encoder and the T5 text encoder to reach consistent responses; 3) \textit{Overlap Loss}. For the semantic ambiguity inherent in similar subjects, we employ an overlap loss to prevent different subjects from being generated in the same location.

Despite these enhancements, there still exists semantic ambiguity when generating multiple similar subjects, as the guidance provided by overlap loss is not explicit enough. Fortunately, this issue can be detected early through cross-attention maps without completing the sampling. Therefore, we propose \textit{Overlap Online Detection and Back-to-Start Sampling Strategy} to address this issue. Specifically, we utilize cross-attention maps to compute each subject's overlap ratio with other subjects when denoising proceeds to a certain early timestep. If all ratios fall below a threshold, sampling proceeds; otherwise, we derive an overlap mask for the subject with the highest ratio, highlighting conflict regions. Then, we restart sampling, applying a \textit{Conflict Mask-Guided Restriction Loss} along with the above losses to prevent that subject from generating in that region. In this way, we explicitly address the coalescence issue of similar semantics. Should overlap persist despite this strategy, the initial noise is deemed a ``bad seed", and we can preemptively reject the sampling to avoid problematic results.

To demonstrate the superiority of the proposed method, we constructed a challenging dataset of similar subjects covering two, three, and four similar subjects, with different categories of animals, plants, fruits, vehicles, balls, \etc. Qualitative and quantitative comparisons as well as user study on this dataset demonstrate the advantages of our method over other state-of-the-art training-free methods in terms of generation quality and success rate. In addition, a full ablation analysis justifies the proposed key components.

To summarize, our contributions are three-fold as below:
\begin{itemize}
	\item We reveal for the first time three core ambiguities existing in MMDiT-based text-to-image models that cause problems in similar subject generation. We hope this will shed light on related generation and editing efforts.

        \item Three novel losses represented by block alignment loss, overlap online detection and back-to-start sampling strategy are proposed to solve the similar subject generation issue.
	
	\item Extensive experiments and analyses are conducted to show the better success rate and generation quality of our method and the necessity of each new design.
\end{itemize}


\section{Related Work}
\label{sec:related}

\noindent\textbf{Text-to-Image Diffusion Models.} After a long period of development~\cite{xu2018attngan, zhang2021cross, kang2023scaling, wei2023hairclipv2, yu2022scaling}, diffusion models~\cite{ho2020denoising, songdenoising} have come to dominate the field of large-scale text-to-image generation. Among them, the representative methods are Stable Diffusion series~\cite{rombach2022high,podell2023sdxl,esser2024scaling}, Imagen~\cite{saharia2022photorealistic}, DALL-E 2~\cite{ramesh2022hierarchical}, \etc. Recently, Stable Diffusion 3~\cite{esser2024scaling} has pushed the quality and resolution of text-to-image synthesis to a new level. In this paper, we intend to explore the boundaries of generative capabilities of this new MMDiT-based text-to-image model and reveal underlying generative mechanisms.

\noindent\textbf{Generation Issues within Diffusion Models.} Even with the enchantment of classical classifier-free guidance~\cite{ho2022classifier, nichol2022glide} or prompt engineering~\cite{liu2022design, witteveen2022investigating, hao2024optimizing, wang2023diffusiondb} techniques, achieving faithful alignment between generated results of text-to-image diffusion models and the given text inputs remains challenging. Thus, in parallel with the development of new generative models through improved modeling paradigms, network architectures, \etc., some work has attempted to mitigate these generative issues present in existing text-to-image models at test time, including: subject neglect~\cite{chefer2023attend,liu2022compositional,fengtraining,zhang2024enhancing,zhang2024object,meral2024conform}, incorrect attribute binding~\cite{rassin2024linguistic,zhang2024object,sueyoshi2024predicated}, incorrect counts~\cite{kang2023counting,binyamin2024make}, incorrect positional control~\cite{chen2024training, dahary2024yourself}, \etc. With the evolution of the diffusion generative model, some common generative issues (\eg, subject neglect, incorrect attribute binding, \etc.) that existed in earlier models~\cite{rombach2022high,podell2023sdxl} have been greatly mitigated in Stable Diffusion 3. However, we observe that subject neglect or mixing is still a difficult issue plaguing SD3 when the input text contains two or especially more similar subjects. Our work focuses on mitigating this problem in a training-free manner based only on the input text without using external signals (\eg, layout masks~\cite{dahary2024yourself}).

\noindent\textbf{Efforts to Alleviate Subject Neglect or Mixing.} As the most typical generative issue, the subject neglect or mixing problem has attracted a lot of work to solve it. In a pioneering effort, Attend-and-Excite~\cite{chefer2023attend} proposes to mitigate it by maximizing the cross-attention activation of the most neglected subject. Based on this, A-Star~\cite{agarwal2023star} further proposes attention segregation loss to reduce the overlap of different subjects on the cross-attention map. Similar to this spirit, Bao~\etal~\cite{bao2024separate} further improve the model's generation ability for new concepts at the expense of fine-tuning the model. CONFORM~\cite{meral2024conform} introduces the concept of contrast learning to address subject neglect, while INITNO~\cite{guo2024initno} highlights the importance of optimizing the initial noise. Recently, Zhang~\etal~\cite{zhang2024object} introduce an object-conditioned energy-based attention map alignment method to alleviate it. Despite efforts by existing approaches, their performance on mitigating similar subject generation issues remains unsatisfactory when applied to MMDiT-based models. This is primarily due to a lack of exploration into the generation mechanisms of the latest MMDiT models. In contrast, our work systematically analyzes the ambiguities that exist in the MMDiT-based model generation process, and specifically designs losses and strategies to mitigate these ambiguities, thus achieving the best performance. 
\section{Proposed Method}
\label{sec:methods}

\subsection{Preliminaries}

\noindent\textbf{Stable Diffusion 3.} As the most cutting-edge text-to-image technology of the latent diffusion model family, SD3~\cite{esser2024scaling} still operates within the latent space of an autoencoder. Generally, an encoder $\mathcal{E}$ is trained to map a given image $x \in \mathcal{X}$ into a spatial latent representation $z = \mathcal{E}(x)$, and a decoder $\mathcal{D}$ is then tasked with reconstructing the input image, ensuring that $\mathcal{D}(\mathcal{E}(x)) \approx x$. 

Discarding the DDPM~\cite{ho2020denoising}, the training objective of SD3 changed to the latest rectified flow~\cite{liuflow}, which connects data and noise in a straight line. Another notable upgrade is that SD3 switches the denoising network from original U-Net~\cite{ronneberger2015u} to the more powerful Multimodal Diffusion Transformer~\cite{peebles2023scalable} (MMDiT) architecture. For a classic SD3-2Billion version, MMDiT consists of $24$ blocks, with one joint self-attention layer in each block accomplishing self-attention and cross-attention for image and text modalities, discussed below.

In the inference phase, a latent variable $z_T$ is sampled from a standard Gaussian distribution $\mathcal{N}(0, 1)$. Then $z_T$ is fed into the MMDiT to perform an iterative denoising process (default $28$ steps) to yield the denoised latent $z_0$. Subsequently, $z_0$ is fed into the decoder $\mathcal{D}$ to obtain the final generated result $x' = \mathcal{D}(z_0)$.

\begin{figure}[t]
	\centering
	\includegraphics[width=0.95\columnwidth]{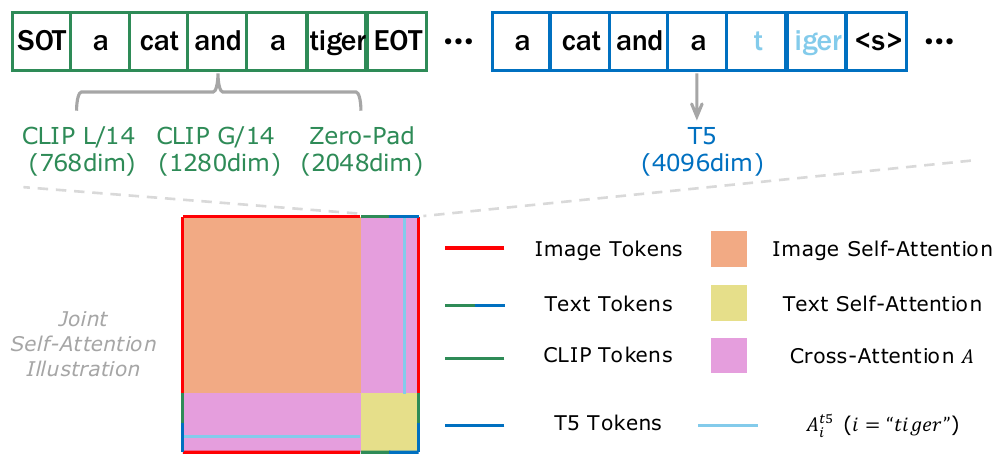} 
	\caption{Illustration for joint self-attention of Stable Diffusion 3.}
	\label{fig:sd3_illus_fig}
\end{figure}

\noindent\textbf{Attention Layers.} Unlike previous versions of SD~\cite{rombach2022high, podell2023sdxl} where the self-attention layer and the cross-attention layer functioned separately, SD3 employs a new information interaction manner, where image and text tokens are concatenated and processed via single self-attention layer. For text conditions, SD3 utilizes three text encoders to provide unprecedented textual contextual information. First, the text embedding from CLIP L/14~\cite{radford2021learning} is concatenated with the embedding from OpenCLIP bigG/14~\cite{cherti2023reproducible} along the channel dimension to form CLIP text tokens. Next, it is concatenated with the embedding from T5-v1.1-XXL~\cite{raffel2020exploring} along the sequence dimension to form the final text tokens.

As shown in Figure~\ref{fig:sd3_illus_fig}, the attention map formed by each self-attention layer of SD3 contains both self-attention of image, self-attention of text, as well as image-text cross-attention and text-image cross-attention. In this work, we take out the last two cross-attention portions and average them to obtain $A$ as our clues to reveal and repair issues present in the SD3 generation process. This is because the cross-attention map $A$ indicates the correlation between the image (\ie, latent hidden representation) and the text condition. We omit the denoising timestep $t$ for all cross-attention map $A$ for the simplicity of representation. Since the sequence dimension of text tokens is composed of concatenated CLIP tokens and T5 tokens, for any subject word $i$ (\eg, ``tiger") in the input text, there exist two corresponding word-based cross-attention maps denoted as $A_{i}^{clip}$ and $A_{i}^{t5}$, respectively. Due to the different tokenizers used in CLIP and T5, a word may consist of multiple tokens. Therefore, $A_{i}^{clip}$ and $A_{i}^{t5}$ are the average of the cross-attention maps of all tokens corresponding to the word. In this study, we focus on operating upon $A_{i}^{clip}$ and $A_{i}^{t5}$ from block $5$ through $12$, as we find that the cross-attention maps from these layers contain the most representative semantic information.

\subsection{Ambiguities Present in MMDiT Generation}

\begin{figure}[t]
	\centering
	\includegraphics[width=0.95\columnwidth]{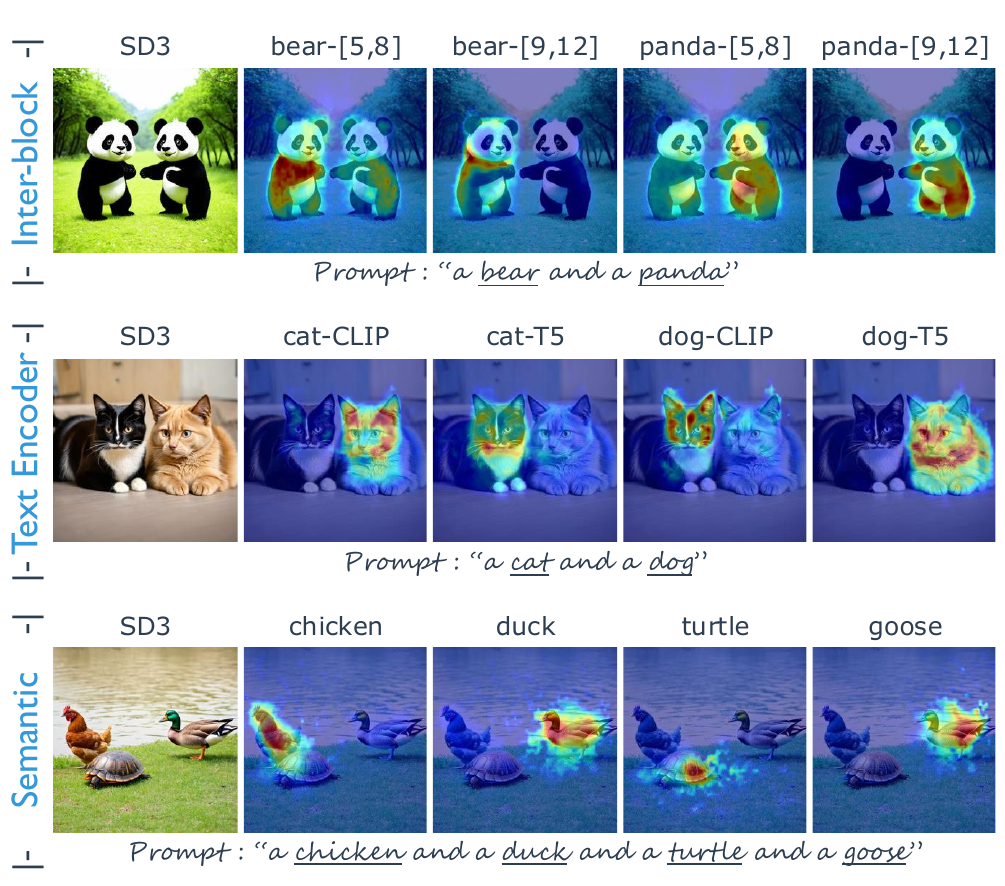} 
	\caption{Three types of ambiguities present in the SD3 generation process, including: inter-block ambiguity, text encoder ambiguity, and semantic ambiguity. All cross-attention maps are from the $5$th step of denoising ($28$ steps in total).}
	\label{fig:ambiguity_fig}
\end{figure}

Thanks to the elaborate designs described above, the generation issues of MMDiT-based text-to-image model have been alleviated tremendously compared to the previous models~\cite{rombach2022high, podell2023sdxl, saharia2022photorealistic}. However, we observe that the subject neglect or mixing problem still troubles the MMDiT model when the input prompt contains two or especially more similar subjects. With the help of subject word based cross-attention maps $A_{i}^{clip}$ and $A_{i}^{t5}$, we try to diagnose the causes of these issues appearing in the MMDiT model generation process. Surprisingly, we discover that in the very first steps of the denoising process, $A_{i}^{clip}$ and $A_{i}^{t5}$ can reveal the causes of the issues that coincide with the final generated results. After a detailed and comprehensive diagnosis, the causes of these issues are categorized into three types of ambiguities that exist in the MMDiT model generation process.

\noindent\textbf{Inter-block Ambiguity.} As revealed by the cross-attention maps, we observe that the blocks of the MMDiT model are progressively clearer in discriminating the semantics of the subjects at each step of the denoising process. Generally, the first four blocks struggle to significantly differentiate between different subject semantics. For regular multi-subject text input (\eg, ``a frog and a bench"), SD3 significantly distinguishes different subjects from the fifth block onwards, thus producing satisfactory results. However, for similar subjects, the $5$th through $8$th blocks become strained. As illustrated in the first row of Figure~\ref{fig:ambiguity_fig}, although the average word-based cross-attention maps for blocks $9$ to $12$ demonstrate that the model has effectively distinguished between ``bear" and ``panda" in these layers, the blocks in layers $5$ to $8$ are still ambiguous. Irretrievably, incorrect semantic information has been injected into intermediate latent hidden representations of the $5$th to $8$th blocks, leading to two pandas being generated. We call this inter-block ambiguity.

\noindent\textbf{Text Encoder Ambiguity.} The introduction of multiple text encoders greatly improves the alignment of the SD3 generated results with the input text. However, there exists ambiguity between text encoders for similar subject generation. The second row of Figure~\ref{fig:ambiguity_fig} shows that CLIP tokens and T5 tokens exhibit a clear ``disagreement" on where ``cat" and ``dog" should be at the $9$th to $12$th blocks, which causes the unfaithful result.

\noindent\textbf{Semantic Ambiguity.} The last row of Figure~\ref{fig:ambiguity_fig} illustrates the semantic ambiguity. When multiple similar subjects are generated, the similarity in shape and appearance of ``duck" and ``goose" causes their attentions are aggregated at same locations, despite the agreement between the text encoders.

\subsection{Mitigating Ambiguities}

\begin{figure*}[t]
	\centering
	\includegraphics[width=\textwidth]{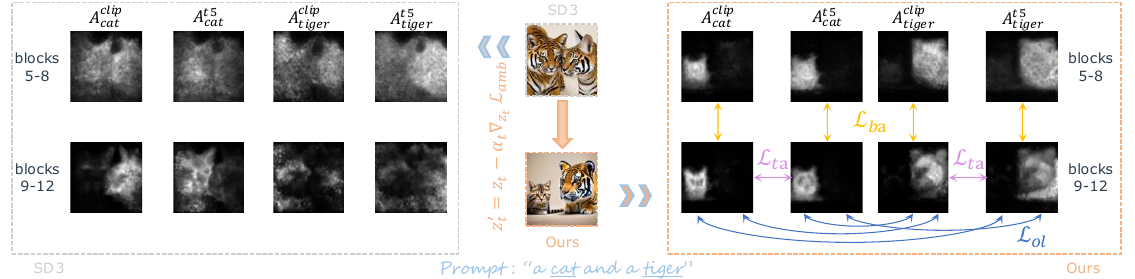}
	\caption{Illustration of three losses for mitigating ambiguities. Left: cross-attention maps from step 5 of Stable Diffusion 3; Right: cross-attention maps from step 5 after imposing losses. Obviously, our cross-attention maps demonstrate strong consistency between blocks and between text encoders with no overlap across subjects.} 
	\label{fig:mitigate_amb}
\end{figure*}

Given these three types of ambiguities, it is tractable to conclude that word-based cross-attention maps of an ideal sampling result should satisfy the following three characteristics: 1) the $5$th-$12$th blocks exhibit strong consistency among themselves; 2) a unified opinion is formed between the CLIP tokens and the T5 tokens; and 3) the overlap between different subjects is as minimal as possible. Therefore, we decide to repair the ambiguous latent on the fly with the hints of word-based cross-attention maps by test-time optimization at early denoising steps. Tailored, three types of losses are proposed as shown in Figure~\ref{fig:mitigate_amb}.

\noindent\textbf{Block Alignment Loss.} Since the discrimination of subjects in MMDiT-based models is progressively clearer as the blocks become deeper, \ie, the blocks in the later layers gradually remove the ambiguities present in the earlier ones, we therefore innovatively propose the block alignment loss $\mathcal{L}_{ba}$ using this mechanism:
\begin{multline}
	\mathcal{L}_{ba}=\frac{1}{2N}\sum^{N}_{i=1}(1-cos(A_{i-[5,8]}^{clip}, A_{i-[9,12]}^{clip}.detach()))\\
 +(1-cos(A_{i-[5,8]}^{t5}, A_{i-[9,12]}^{t5}.detach())),
        \label{eq:block_align}
\end{multline}
where $N$ represents the number of subject words present in the input text, $cos$ denotes cosine similarity, and $A_{i-[5,8]}^{clip}$ stands for the average of cross-attention maps from the $5$th to the $8$th blocks corresponding to the CLIP tokens of subject word $i$. Intuitively, we detach the more ``correct" CLIP and T5 cross-attention maps from the $9$th to the $12$th blocks as ground truth to constrain the $5$th to the $8$th blocks to be consistent with them. In this way, we delicately exploit the self-refinement mechanism present in the transformer structure of MMDiT to mitigate the semantic leakage of the previous blocks caused by ambiguity.

\noindent\textbf{Text Encoder Alignment Loss.} Against text encoder ambiguity, unlike inter-block ambiguity, it is difficult to conclude whose activation is more reliable between CLIP text encoder and T5 text encoder. Therefore, we implicitly constrain both to form a unified opinion:
\begin{equation}
	\mathcal{L}_{ta}=\frac{1}{N}\sum^{N}_{i=1}(1-cos(A_{i-[9,12]}^{clip}, A_{i-[9,12]}^{t5})).
\end{equation}
\noindent\textbf{Overlap Loss.} To alleviate semantic ambiguity, we leverage overlap loss $\mathcal{L}_{ol}$ to mitigate overlap between different subjects as follows:
\begin{multline}
	\mathcal{L}_{ol}=\sum_{1 \le i < j \le N} \hat{A}_{i-[9,12]}^{clip} \cdot \hat{A}_{j-[9,12]}^{clip}+\hat{A}_{i-[9,12]}^{t5} \cdot \hat{A}_{j-[9,12]}^{t5}\\
 +\hat{A}_{i-[9,12]}^{clip} \cdot \hat{A}_{j-[9,12]}^{t5}+\hat{A}_{i-[9,12]}^{t5} \cdot \hat{A}_{j-[9,12]}^{clip},
\end{multline}
where $i$ and $j$ represent two different subject words and $\hat{A}$ represents the normalized attention map.

In summary, the loss function $\mathcal{L}_{amb}$ for these three types of ambiguities is defined as follows:
\begin{equation}
\mathcal{L}_{amb}=\lambda_{ba}\mathcal{L}_{ba} + \lambda_{ta}\mathcal{L}_{ta}+ \lambda_{ol}\mathcal{L}_{ol},
\end{equation}
where $ \lambda_{ba} $, $ \lambda_{ta} $, $ \lambda_{ol} $ are set to $ 1 $, $ 0.2 $, $ 0.001 $ respectively depending on their magnitude and importance. Based on the directions provided by $\mathcal{L}_{amb}$, we shift the latent representation at early stage of the denoising process. At each early timestep $t$, the latent representation $z_t$ is updated as follows: $z_{t}^{'}=z_t - \alpha_{t}\nabla_{z_t}\mathcal{L}_{amb}$.

\subsection{Advanced Strategies for Semantic Ambiguity}

\begin{figure}[t]
	\centering
	\includegraphics[width=0.96\columnwidth]{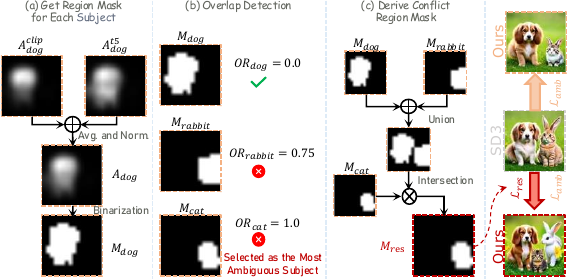} 
	\caption{Overview of overlap online detection and back-to-start sampling strategy. Prompt: ``\textit{a \underline{dog} and a \underline{cat} and a \underline{rabbit}}".}
	\label{fig:semantic_fig}
\end{figure}

Thanks to this on-the-fly repair technique, the generation issues are greatly mitigated. However, we observe that for the case of three or more similar subjects generated, the problem of semantic ambiguity still exists to some extent despite the overlap loss imposed. We consider the underlying reason to be that overlap loss is implicit and does not provide more direct and effective guidance. It only penalizes overlap between subjects but fails to provide more effective information (\eg, where subjects should or should not be generated), leading to the situation that there are still semantic information of similar subjects injected into the same location. Thankfully, we can diagnose and further repair the problem at the early stage of denoising with the hint of word-based cross-attention maps. The novel \textit{Overlap Online Detection and Back-to-Start Sampling Strategy} is proposed as follows.

\noindent\textbf{Overlap Online Detection.} When the denoising process proceeds to step $5$, we evaluate the repair effect brought by applying $\mathcal{L}_{amb}$ in the previous steps. As shown in Figure~\ref{fig:semantic_fig}-(a), for each subject word $i$, we compute the mean of its corresponding cross-attention maps of CLIP tokens and T5 tokens and normalize it: $A_{i}=Norm(\frac{1}{2}A_{i-[9,12]}^{clip}+\frac{1}{2}A_{i-[9,12]}^{t5}), \quad i = 1, 2, \dots, N$. In this way, we take full advantage of the semantics from the CLIP text encoder and the T5 text encoder to identify the location of the subject, while also absorbing the text encoder ambiguity into the evaluation. Then, we binarize $A_{i}$ pixel-by-pixel with a threshold of $0.2$ thereby derive the region mask $M_{i}$ corresponding to each subject: $M_i = \mathbb{I}(A_i > 0.2), \quad i = 1, 2, \dots, N$. Finally, the overlap ratio $OR_{i}$ of each subject with other subjects is calculated as follows:
\begin{equation}
	OR_i = \frac{\sum_{x, y} \left( M_i(x, y) \cap \bigcup_{j \neq i} M_j(x, y) \right)}{\sum_{x, y} M_i(x, y)},
\end{equation}
where $\bigcup_{j \neq i} M_j(x, y)$ denotes the union set of the remaining subject masks, which is then used to compute the intersection set with the mask of subject $i$, and $\sum_{x, y}$ denotes summation over all pixel positions. After obtaining overlap ratios of all subjects, we use $0.2$ as a threshold for detection (Figure~\ref{fig:semantic_fig}-(b)). If the overlap ratios of all subjects are below $0.2$, it indicates that everything is normal and sampling continues. If there are subjects with an overlap ratio greater than $0.2$, it implies that subject neglect or mixing is about to occur. At this point, the following strategy is triggered.

\noindent\textbf{Back-to-Start Sampling.} First, we identify the subject with the largest overlap ratio: $i^* = \arg\max_{i} OR_i, \quad i = 1, 2, \dots, N$. Then, the conflict region mask is derived: $M_{res}=M_{i^{*}}(x, y) \cap \bigcup_{j \neq i^*} M_j(x, y)$ as shown in Figure~\ref{fig:semantic_fig}-(c). Finally, with this conflict region mask, we set the sampling back to the start. Still using the original initial noise as the starting point, an additional \textit{Conflict Mask-Guided Restriction Loss} is applied to the most ambiguous subject $i^*$ on top of the original three losses:
\begin{multline}
    \mathcal{L}_{res}=\frac{1}{2}(\frac{\sum_{x, y} (A_{i-[5,12]}^{clip}(x, y) \cdot M_{res}(x, y))}{\sum_{x, y} A_{i-[5,12]}^{clip}(x, y)} +\\
    \frac{\sum_{x, y} (A_{i-[5,12]}^{t5}(x, y) \cdot M_{res}(x, y))}{\sum_{x, y} A_{i-[5,12]}^{t5}(x, y)}) \cdot \mathbb{I}(i = i^*).
\end{multline}
Here, our philosophy is to identify the most ambiguous subject and constrain it from appearing in its original overlapping position from the start of sampling, but we do not explicitly specify where it should appear. By doing so, we effectively avoid the overlapping problem of similar semantic subjects while largely preserving the freedom of generation and diversity of results. At this stage, the total loss function becomes $\mathcal{L}=\mathcal{L}_{amb}+\mathcal{L}_{res}$, while the latent representation $z_t$ is updated as follows: $z_{t}^{'}=z_t - \alpha_{t}\nabla_{z_t}\mathcal{L}$.

\noindent\textbf{Reject Sampling.} When back-to-start sampling strategy proceeds to step $5$, we perform another overlap online detection. If we still find an overlap ratio greater than threshold, we consider that even an additional attempt still fails to repair the ``bad" seed, and therefore reject this sampling.
\section{Experiments}
\label{sec:exp}

\noindent\textbf{Implementation Details.} We deploy our approach on Stable Diffusion $3$ Medium ($2$B). Following official recommendations, the guideline scale is set to $7.0$, and the number of denoising steps is set to $28$ by default. The latent update scale $\alpha_{t}$ is set to $30$. Consistent with previous methods~\cite{chefer2023attend, meral2024conform}, latent is optimized only in the first half of the denoising steps (\ie, steps $1$-$14$). Differently, they choose to optimize the initial noise for multiple iterations, which strengthens repair effect but also causes initial noise to deviate from the Gaussian distribution resulting in undesirable synthesis quality. Therefore, we choose to perform multiple iterations of optimization in the early stages of denoising (steps $3$-$5$) to trade-off between the two. All experiments were performed on single $48$G NVIDIA A6000 GPU.

\subsection{Quantitative and Qualitative Comparison}

\begin{figure*}[t]
	\begin{center}
		\setlength{\tabcolsep}{0.5pt}
		\begin{tabular}{m{0.3cm}<{\centering}m{1.68cm}<{\centering}m{1.68cm}<{\centering}m{1.68cm}<{\centering}m{1.68cm}<{\centering}m{1.68cm}<{\centering}m{1.68cm}<{\centering}m{1.68cm}<{\centering}m{1.68cm}<{\centering}m{1.68cm}<{\centering}m{1.68cm}<{\centering}}
			 & \multicolumn{2}{c}{\scriptsize{\textit{``a \underline{rose} and a \underline{tulip}"}}} & \multicolumn{2}{c}{\scriptsize{\makecell{\scriptsize{\textit{``a \underline{bike} and a \underline{scooter}}} \\ \scriptsize{\textit{and a \underline{skateboard}"}}}}} & \multicolumn{2}{c}{\scriptsize{\makecell{\scriptsize{\textit{``a \underline{basketball} and a \underline{volleyball}}} \\ \scriptsize{\textit{and a \underline{baseball}"}}}}} & \multicolumn{2}{c}{\scriptsize{\makecell{\scriptsize{\textit{``a \underline{lemon} and an \underline{apple} and}} \\ \scriptsize{\textit{an \underline{orange} and a \underline{strawberry}"}}}}} & \multicolumn{2}{c}{\scriptsize{\makecell{\scriptsize{\textit{``a \underline{cat} and a \underline{dog} and}} \\ \scriptsize{\textit{a \underline{parrot} and a \underline{hamster}"}}}}}
			\\
   
			\raisebox{0.15cm}{\rotatebox[origin=c]{90}{\footnotesize{{SD3}}}}
			&\includegraphics[width=1.65cm]{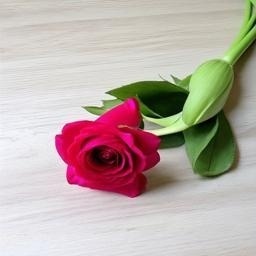}
			&\includegraphics[width=1.65cm]{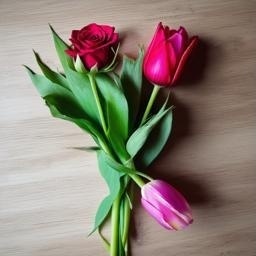}
			&\includegraphics[width=1.65cm]{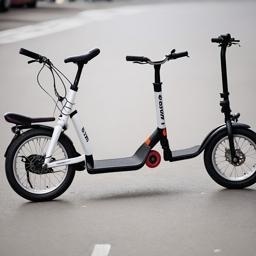}
			&\includegraphics[width=1.65cm]{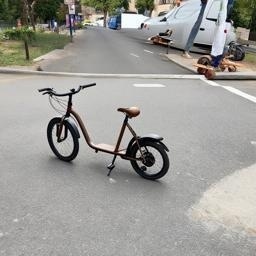}
			&\includegraphics[width=1.65cm]{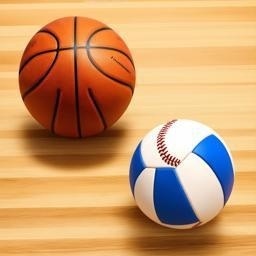}
			&\includegraphics[width=1.65cm]{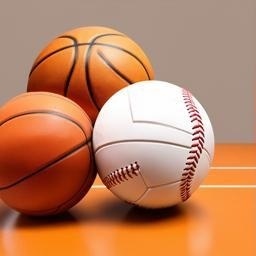}
			&\includegraphics[width=1.65cm]{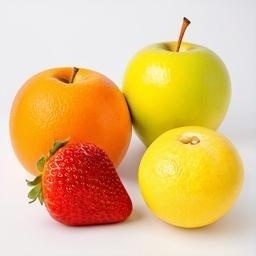}
			&\includegraphics[width=1.65cm]{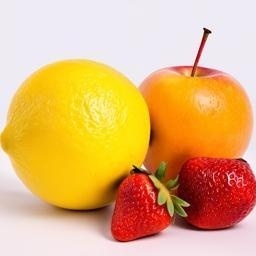}
			&\includegraphics[width=1.65cm]{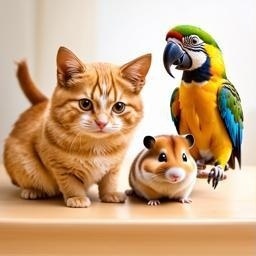}
			&\includegraphics[width=1.65cm]{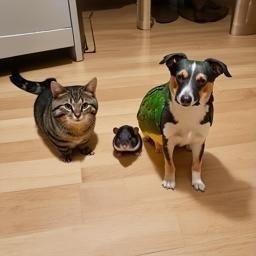}
			\\

			\raisebox{0.20cm}{\rotatebox[origin=c]{90}{\footnotesize{{A\&E}}}}
			&\includegraphics[width=1.65cm]{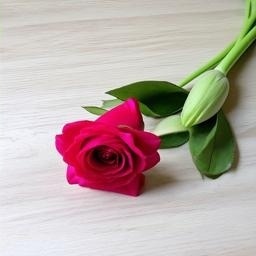}
			&\includegraphics[width=1.65cm]{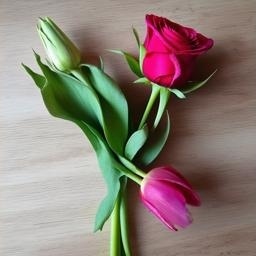}
			&\includegraphics[width=1.65cm]{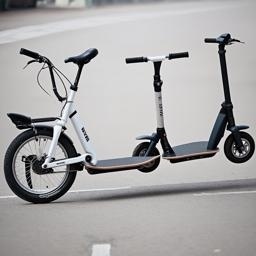}
			&\includegraphics[width=1.65cm]{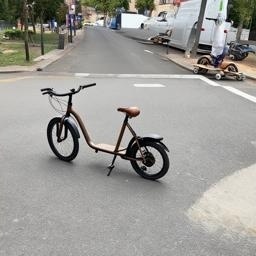}
			&\includegraphics[width=1.65cm]{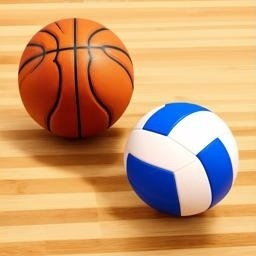}
			&\includegraphics[width=1.65cm]{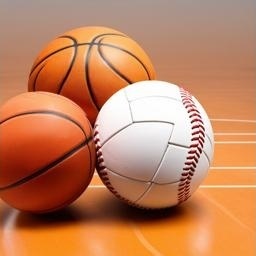}
			&\includegraphics[width=1.65cm]{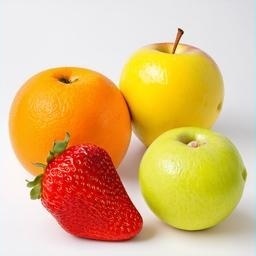}
			&\includegraphics[width=1.65cm]{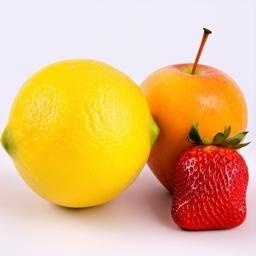}
			&\includegraphics[width=1.65cm]{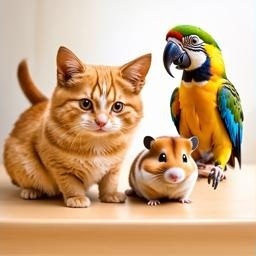}
			&\includegraphics[width=1.65cm]{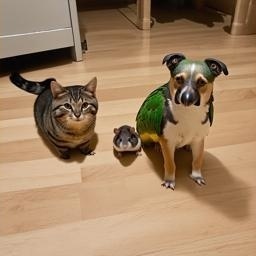}
			\\

			\raisebox{0.35cm}{\rotatebox[origin=c]{90}{\footnotesize{{EBAMA}}}}
			&\includegraphics[width=1.65cm]{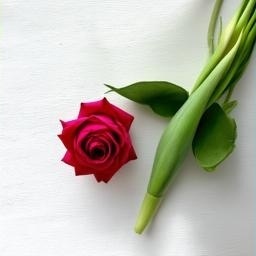}
			&\includegraphics[width=1.65cm]{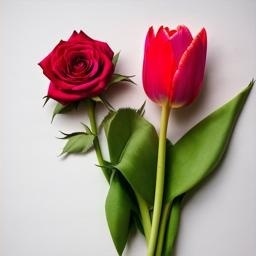}
			&\includegraphics[width=1.65cm]{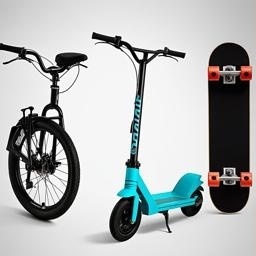}
			&\includegraphics[width=1.65cm]{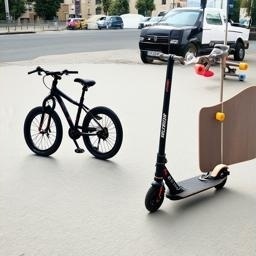}
			&\includegraphics[width=1.65cm]{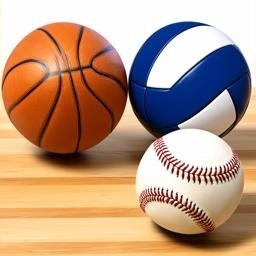}
			&\includegraphics[width=1.65cm]{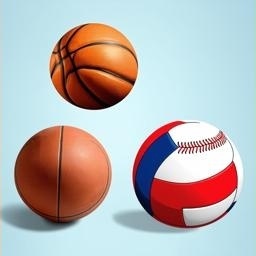}
			&\includegraphics[width=1.65cm]{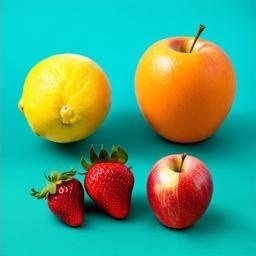}
			&\includegraphics[width=1.65cm]{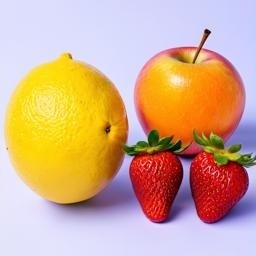}
			&\includegraphics[width=1.65cm]{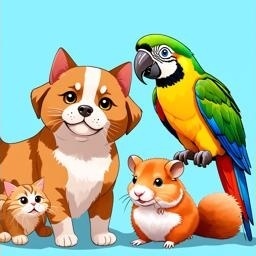}
			&\includegraphics[width=1.65cm]{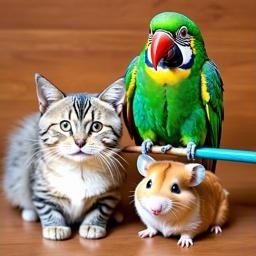}
			\\

			\raisebox{0.55cm}{\rotatebox[origin=c]{90}{\footnotesize{{CONFORM}}}}
			&\includegraphics[width=1.65cm]{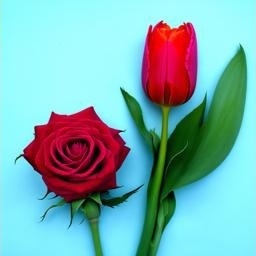}
			&\includegraphics[width=1.65cm]{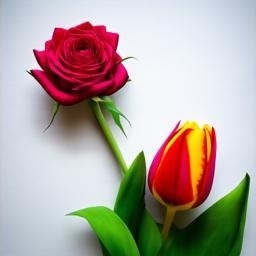}
			&\includegraphics[width=1.65cm]{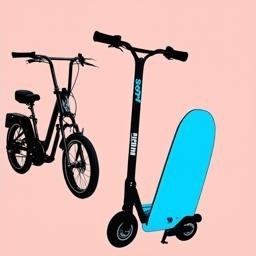}
			&\includegraphics[width=1.65cm]{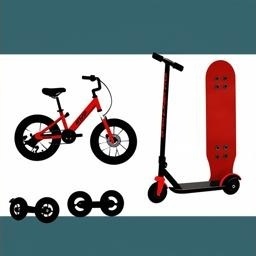}
			&\includegraphics[width=1.65cm]{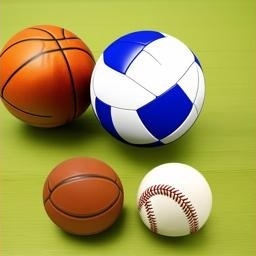}
			&\includegraphics[width=1.65cm]{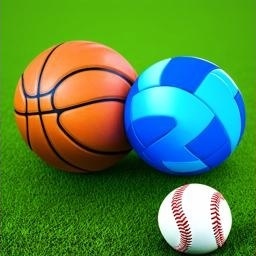}
			&\includegraphics[width=1.65cm]{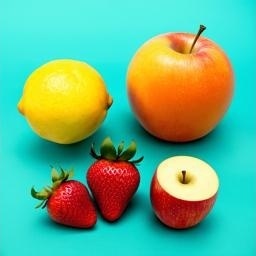}
			&\includegraphics[width=1.65cm]{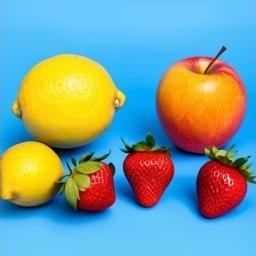}
			&\includegraphics[width=1.65cm]{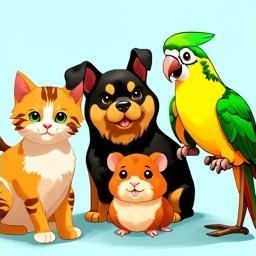}
			&\includegraphics[width=1.65cm]{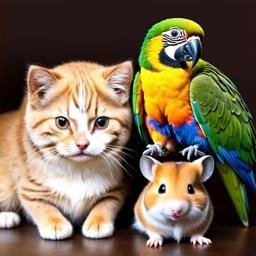}
			\\

			\raisebox{0.20cm}{\rotatebox[origin=c]{90}{\footnotesize{{Ours}}}}
			&\includegraphics[width=1.65cm]{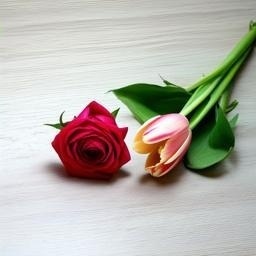}
			&\includegraphics[width=1.65cm]{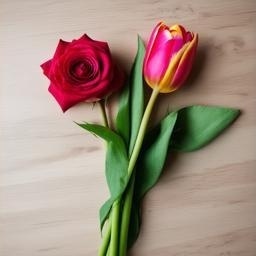}
			&\includegraphics[width=1.65cm]{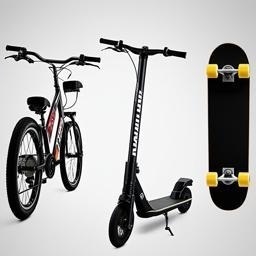}
			&\includegraphics[width=1.65cm]{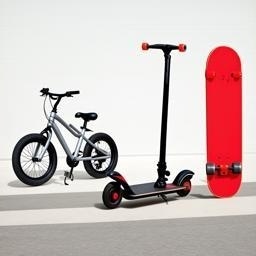}
			&\includegraphics[width=1.65cm]{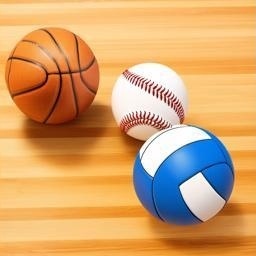}
			&\includegraphics[width=1.65cm]{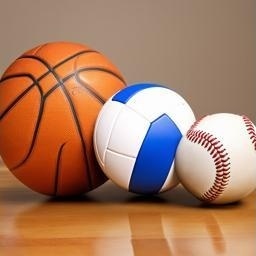}
			&\includegraphics[width=1.65cm]{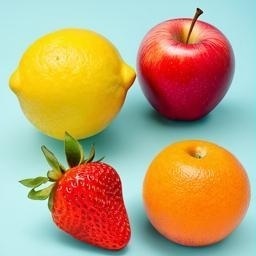}
			&\includegraphics[width=1.65cm]{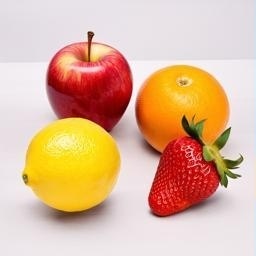}
			&\includegraphics[width=1.65cm]{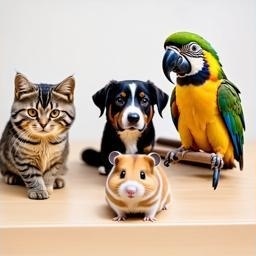}
			&\includegraphics[width=1.65cm]{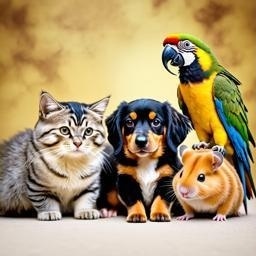}
			\\
			
		\end{tabular}
	\end{center}
	\caption{Qualitative comparison with A\&E~\cite{chefer2023attend}, EBAMA~\cite{zhang2024object}, CONFORM~\cite{meral2024conform} on prompts containing two, three, and four similar subjects. Our approach mitigates the subject neglect or mixing problems present in SD3~\cite{esser2024scaling} while maintaining the best generation quality.}
	\label{fig:quail_comparsion}
\end{figure*}

\noindent\textbf{Dataset.} As stated before, the existing datasets~\cite{chefer2023attend} are no longer difficult for the latest state-of-the-art models like SD3. To demonstrate the superiority of the proposed method, we constructed a challenging similar-subject text description dataset with the help of ChatGPT~\cite{openai2024chatgpt}, which covers different categories such as animals, plants, fruits, vehicles, balls, \etc. We generated $30$ prompts (\eg, ``a [subjectA] and a [subjectB]") each for descriptions with two similar subjects, three similar subjects, and four similar subjects, resulting in a total of $90$ prompts. It is noteworthy that this construction is only for the convenience of quantitative evaluation and illustration, and our method can be applied to various complex prompts as demonstrated in Figure~\ref{fig:teaser}.

\noindent\textbf{Qualitative Comparison.} We compare proposed method with the following state-of-the-art methods of test-time optimization: A\&E~\cite{chefer2023attend}, CONFORM~\cite{meral2024conform}, EBAMA~\cite{zhang2024object}. All of these methods were reproduced on the SD3 according to their official codes. Qualitative comparisons are provided in Figure~\ref{fig:quail_comparsion}. The strategy proposed by A\&E to maximize cross attention activation of the most neglected subject struggles to provide effective guidance in more challenging scenarios with multiple similar subjects, leading to its lowest repair success rate. The approaches proposed by EBAMA and CONFORM share a somewhat similar philosophy to our overlap loss, and thus they demonstrate some success, despite some artifacts (\eg, ``a bike and a scooter and a skateboard"). In addition, CONFORM over-optimizes the initial noise, which mitigates the generation problem but causes it to deviate from the original Gaussian distribution, deteriorating the quality of the final synthesis result (\eg, the first example of ``a cat and a dog and a parrot and a hamster" is transformed into a cartoon image).  Thanks to our block alignment loss, text encoder alignment loss, tailored to address ambiguities in SD3, and overlap online detection and back-to-start sampling strategy proposed to tackle semantic ambiguity of multiple similar subjects, our approach effectively mitigates the subject neglect or mixing problem while maintaining the best generation quality.

\noindent\textbf{Quantitative Comparison.} For each of these $90$ prompts, we generate $100$ images using $100$ random seeds applied across all methods. The quantitative evaluation metrics consist of: SR (G DINO), SR (GPT 4o), FID. SR (G DINO) and SR (GPT 4o) represent the Success Rate (whether or not the subjects all appear in the results) calculated using the state-of-the-art open-set object detection method Grounding DINO~\cite{liu2024grounding} and the most powerful and efficient intelligent model GPT 4o-mini~\cite{gpt4o_mini}, respectively. The implementation details of these two success rates are provided in the supplementary material. Balancing cost overhead and accuracy, for SR (GPT 4o) we randomly selected $50$ results per prompt ($4500$ in total) for each method to call GPT 4o-mini API evaluation, which does not affect the statistical accuracy. In terms of FID~\cite{Heusel2017GANsTB}, we consider the original SD3 sampling results as an upper bound for the synthesis quality. Therefore, FID~\cite{Heusel2017GANsTB} is calculated between the different methods and the sampling results of SD3, representing the ability of different methods to maintain the original generation quality of SD3. The quantitative results are provided in Table~\ref{tab:quant_comparsion}. Across settings of two, three, and four similar subjects, our method achieves the best success rate, as double-validated by Grounding DINO and GPT 4o-mini. In terms of FID, the method of A\&E~\cite{chefer2023attend} struggles to provide effective guidance for similar subject scenarios, making its results closer to the original SD3 sampling results (as evidenced in Fig.~\ref{fig:quail_comparsion}), which leads to its optimal FID, although it is meaningless. After disregarding it, our method is basically the best in terms of FID metric, which proves that our method achieves the best repair success rate while still maintaining its generation quality to the maximum extent. In two similar subjects setting, EBAMA~\cite{zhang2024object} and CONFORM~\cite{meral2024conform} also achieve a good repair success rate. When it comes to the three and four scenarios, the advantages of our method become more obvious, which is consistent with the visual comparisons in Fig.~\ref{fig:quail_comparsion}. In addition, the proposed reject sampling strategy filters out the failing examples without any additional overhead thus further improving the success rate, especially for the four similar subjects setting.

\begin{table*}[t]
	\centering
	\small
	\setlength{\tabcolsep}{2.1pt}{
		\begin{tabular}{l|lll|lll|lll}
			\hline
			\multicolumn{1}{ c }{{}}	& \multicolumn{3}{ c }{{\small{Two Similar Subjects}}} & \multicolumn{3}{ c }{{\small{Three Similar Subjects}}} & \multicolumn{3}{ c }{{\small{Four Similar Subjects}}} \\
			\hline
			\small{Methods} & \small{SR (G DINO) $\uparrow$} & \small{SR (GPT 4o) $\uparrow$} & \small{FID $\downarrow$} & \small{SR (G DINO) $\uparrow$} & \small{SR (GPT 4o) $\uparrow$} & \small{FID $\downarrow$} & \small{SR (G DINO) $\uparrow$} & \small{SR (GPT 4o) $\uparrow$} & \small{FID $\downarrow$} \\
			\hline
			SD3  & \small{53.1\%} & \small{68.2\%} & \small{-} & \small{37.9\%} & \small{31.2\%} & \small{-} & \small{33.1\%} & \small{22.5\%} & \small{-} \\
			A\&E  & \small{54.8\% (\textcolor{blue}{+1.7})} & \small{69.9\% (\textcolor{blue}{+1.7})} & \small{4.15} & \small{41.2\% (\textcolor{blue}{+3.3})} & \small{33.8\% (\textcolor{blue}{+2.6})} & \small{4.25} & \small{34.0\% (\textcolor{blue}{+0.9})} & \small{24.2\% (\textcolor{blue}{+1.7})} & \small{3.89} \\
			EBAMA  & \small{58.8\% (\textcolor{blue}{+5.7})} & \small{89.1\% (\textcolor{blue}{+20.9})} & \small{14.51} & \small{49.1\% (\textcolor{blue}{+11.2})} & \small{55.9\% (\textcolor{blue}{+24.7})} & \small{16.40} & \small{41.0\% (\textcolor{blue}{+7.9})} & \small{37.7\% (\textcolor{blue}{+15.2})} & \small{17.69} \\
			CONFORM  & \small{61.9\% (\textcolor{blue}{+8.8})} & \small{89.5\% (\textcolor{blue}{+21.3})} & \small{27.40} & \small{57.1\% (\textcolor{blue}{+19.2})} & \small{72.1\% (\textcolor{blue}{+40.9})} & \small{35.55} & \small{42.4\% (\textcolor{blue}{+9.3})} & \small{43.7\% (\textcolor{blue}{+21.2})} & \small{30.86} \\
			Ours-(w/o RS)  & \small{\textbf{68.8\%} (\textcolor{blue}{+15.7})} & \small{\textbf{90.9\%} (\textcolor{blue}{+22.7})} & \small{11.80} & \small{\textbf{66.7\%} (\textcolor{blue}{+28.8})} & \small{\textbf{80.2\%} (\textcolor{blue}{+49.0})} & \small{14.26} & \small{\textbf{52.7\%} (\textcolor{blue}{+19.6})} & \small{\textbf{58.5\%} (\textcolor{blue}{+36.0})} & \small{18.17} \\
			Ours-(w/ RS)  & \small{\textbf{68.8\%} (\textcolor{blue}{+15.7})} & \small{\textbf{90.9\%} (\textcolor{blue}{+22.7})} & \small{11.80} & \small{\textbf{67.8\%} (\textcolor{blue}{+29.9})} & \small{\textbf{83.9\%} (\textcolor{blue}{+52.7})} & \small{14.60} & \small{\textbf{60.8\%} (\textcolor{blue}{+27.7})} & \small{\textbf{68.7\%} (\textcolor{blue}{+46.2})} & \small{18.98} \\
                \hline
		\end{tabular}
	}
	\caption{Quantitative comparison with A\&E~\cite{chefer2023attend}, EBAMA~\cite{zhang2024object}, CONFORM~\cite{meral2024conform}. SR (G DINO) and SR (GPT 4o) represent the Success Rate (whether or not the subjects all appear in the results) calculated using Grounding DINO~\cite{liu2024grounding} and GPT 4o-mini~\cite{gpt4o_mini}, respectively. Blue numbers indicate the percent gain of different methods against SD3. FID~\cite{Heusel2017GANsTB} is calculated between the different methods and the sampling results of SD3, representing the ability of different methods to maintain the original generation quality of SD3. Ours-(w/ RS) and Ours-(w/o RS) represent whether or not our method is applied with the Reject Sampling strategy, respectively.}
	\label{tab:quant_comparsion}
\end{table*}

\noindent\textbf{User Study.} Further, we conducted a user study involving $32$ participants with computer vision research backgrounds. Each participant received $25$ groups of randomized results, in which two, three, and four similar subjects accounted for $20\%$, $40\%$, and $40\%$, respectively. Participants were asked to choose the best option considering the alignment of generated images with texts and the visual quality of the images. User preference rates are provided in Table~\ref{tab:user_study}, where our method is greatly outperforming other counterparts.

\begin{table}[h]
    \centering
    \begin{tabular}{lccccc}
        \hline
        \small{Methods} & \small{SD3} &\small{~\cite{chefer2023attend}}&\small{~\cite{zhang2024object}}&\small{~\cite{meral2024conform}}& \small{Ours} \\
        \hline
        \small{Preference} & \small{1.63\%} & \small{1.88\%} & \small{13.25\%} & \small{9.88\%} & \textbf{\small{73.38\%}} \\
        \hline
    \end{tabular}
    \caption{Results from a user study with $32$ participants.}
    \label{tab:user_study}
\end{table}

\subsection{Ablation Analysis}

\begin{figure}[t]
	\centering
	\includegraphics[width=0.95\columnwidth]{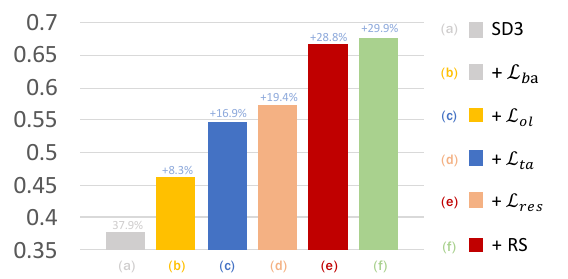} 
	\caption{Quantitative ablation on proposed losses and strategies. We evaluate the gain of different components using SR (G DINO) metric on three similar subjects dataset. From (a) to (f), each setting adds an additional component to the previous one. Blue numbers indicate the percent gain of different settings against SD3.}
	\label{fig:ablation_loss}
\end{figure}

\noindent\textbf{Quantitative Evaluation of the Importance of Different Losses and Strategies.} We perform ablation analysis using SR (G DINO) metric on the most representative three similar subjects dataset to evaluate the gain of proposed different losses and strategies. As shown in Figure~\ref{fig:ablation_loss}, block alignment loss $\mathcal{L}_{ba}$ and overlap loss $\mathcal{L}_{ol}$ bring additional success rate improvements of $8.3\%$ and $8.6\%$, respectively. The $\mathcal{L}_{amb}$ (setting (d)) formed after combining text encoder alignment loss $\mathcal{L}_{ta}$ boosts the success rate to $57.3\%$. After applying overlap online detection and back-to-start sampling strategy (setting (e)) on top of this, the success rate is considerably boosted to $66.7\%$. Moreover, reject sampling strategy (setting (f)) further improves the success rate without extra overhead. Below, we further provide visual ablation results for the two core contributions of this paper.

\begin{figure}[h]
	\begin{center}
		\setlength{\tabcolsep}{0.5pt}
		\begin{tabular}{cccccc}
			\small{SD3} & \small{+$\mathcal{L}_{ba}^{rev}$} & \small{+$\mathcal{L}_{ba}$} & \small{SD3} & \small{+$\mathcal{L}_{ba}^{rev}$} & \small{+$\mathcal{L}_{ba}$}
			\\
			\includegraphics[width=1.34cm]{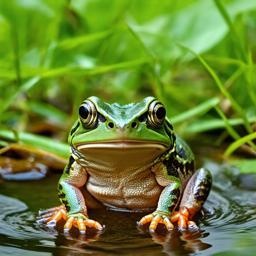}
			&\includegraphics[width=1.34cm]{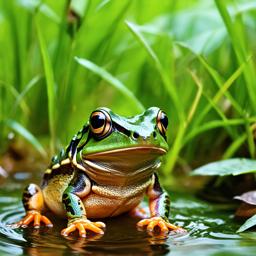}
			&\includegraphics[width=1.34cm]{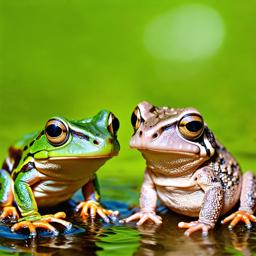}
			&\includegraphics[width=1.34cm]{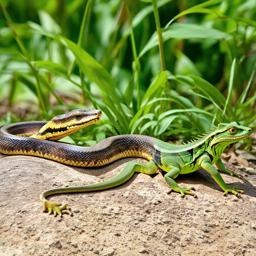}
			&\includegraphics[width=1.34cm]{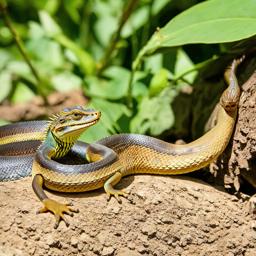}
			&\includegraphics[width=1.34cm]{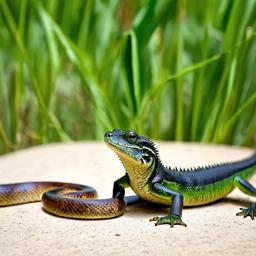}
			\\

			\includegraphics[width=1.34cm]{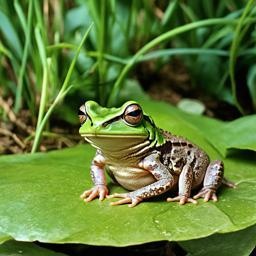}
			&\includegraphics[width=1.34cm]{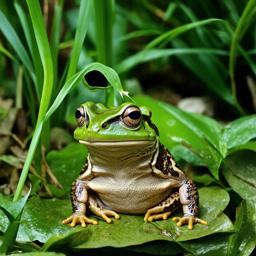}
			&\includegraphics[width=1.34cm]{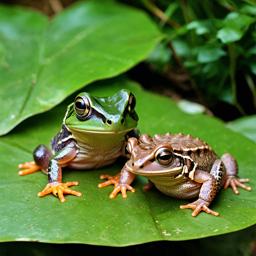}
			&\includegraphics[width=1.34cm]{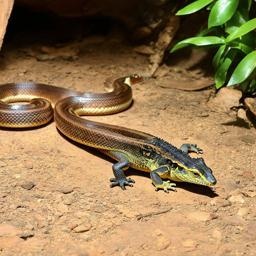}
			&\includegraphics[width=1.34cm]{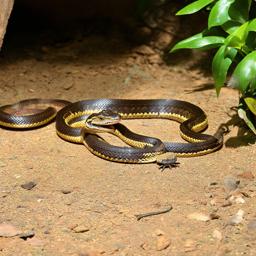}
			&\includegraphics[width=1.34cm]{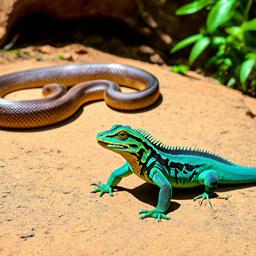}
			\\
   
                \multicolumn{3}{c}{\small{\textit{``a \underline{frog} and a \underline{toad}"}}} & \multicolumn{3}{c}{\small{\textit{``a \underline{snake} and a \underline{lizard}"}}} \\
		\end{tabular}
	\end{center}
	\caption{Qualitative ablation on block alignment loss $\mathcal{L}_{ba}$.}
	\label{fig:ablation_ba}
\end{figure}

\noindent\textbf{Superiority of Block Alignment Loss.} To verify the superiority of our block alignment loss design, we rewrite $\mathcal{L}_{ba}$ as $\mathcal{L}_{ba}^{rev}$. Specifically, $\mathcal{L}_{ba}^{rev}$ reverses the constraint relationship in the original Eq.~\ref{eq:block_align}, \ie, the cross-attention map from blocks 5-8 is detached as the ground truth constraining blocks 9-12 to be consistent with it. Visual comparisons are shown in Figure~\ref{fig:ablation_ba}. Applying $\mathcal{L}_{ba}^{rev}$ not only does not alleviate the generation problem but also exacerbates it to some extent, while applying our $\mathcal{L}_{ba}$ effectively alleviates the inter-block ambiguity. This justifies the correctness of our discovery that the discrimination of subjects within the MMDiT architecture becomes progressively clearer as the block deepens and the rationality of our design that skillfully exploits this feature to mitigate inter-block ambiguity in the self-refinement manner.

\begin{figure}[h]
	\begin{center}
		\setlength{\tabcolsep}{0.5pt}
		\begin{tabular}{cccccc}
			\small{SD3} & \small{+$\mathcal{L}_{amb}$} & \small{+$\mathcal{L}$} & \small{SD3} & \small{+$\mathcal{L}_{amb}$} & \small{+$\mathcal{L}$}
			\\
			\includegraphics[width=1.34cm]{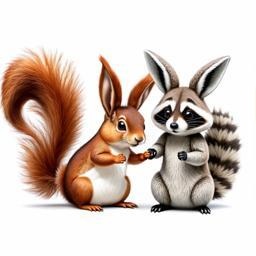}
			&\includegraphics[width=1.34cm]{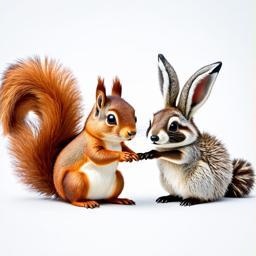}
			&\includegraphics[width=1.34cm]{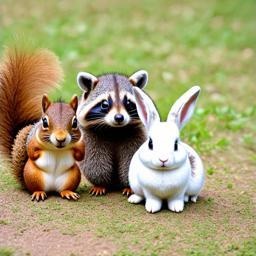}
			&\includegraphics[width=1.34cm]{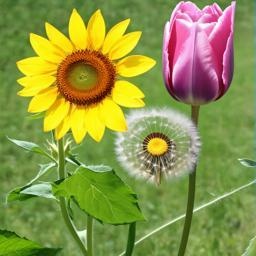}
			&\includegraphics[width=1.34cm]{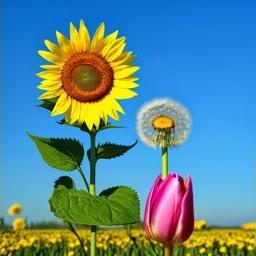}
			&\includegraphics[width=1.34cm]{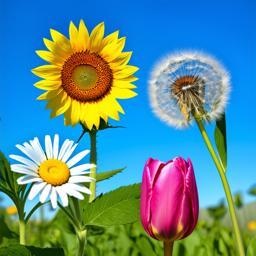}
			\\

			\includegraphics[width=1.34cm]{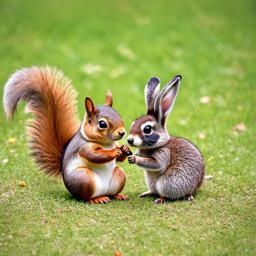}
			&\includegraphics[width=1.34cm]{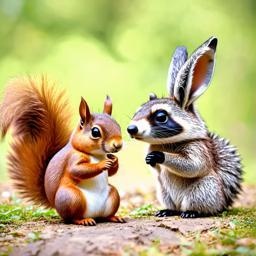}
			&\includegraphics[width=1.34cm]{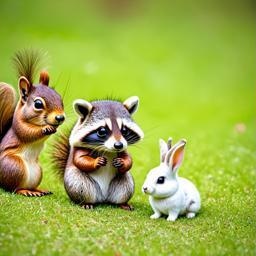}
			&\includegraphics[width=1.34cm]{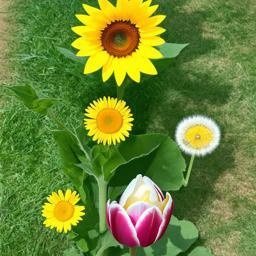}
			&\includegraphics[width=1.34cm]{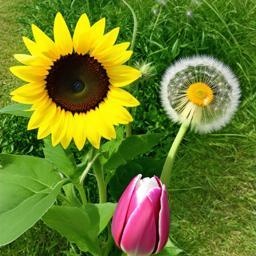}
			&\includegraphics[width=1.34cm]{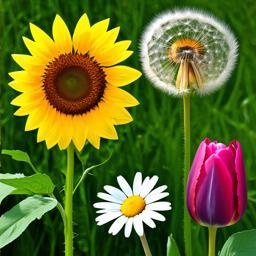}
			\\

                \multicolumn{3}{c}{\small{\makecell{\small{\textit{``a \underline{squirrel} and a \underline{rabbit}}} \\ \small{\textit{and a \underline{raccoon}"}}}}} & \multicolumn{3}{c}{\small{\makecell{\small{\textit{``a \underline{sunflower} and a \underline{daisy} and}} \\ \small{\textit{a \underline{dandelion} and a \underline{tulip}"}}}}} \\
		\end{tabular}
	\end{center}
	\caption{Qualitative ablation on overlap online detection and back-to-start sampling strategy. $\mathcal{L}=\mathcal{L}_{amb}+\mathcal{L}_{res}$.}
	\label{fig:ablation_restrict}
\end{figure}

\noindent\textbf{Effectiveness of Overlap Online Detection and Back-to-Start Sampling Strategy.} As shown in Figure~\ref{fig:ablation_restrict}, despite the overlap loss included in $\mathcal{L}_{amb}$, the guidance it provides is not direct enough, leading to a decrease in subject neglect or mixing mitigation performance for cases with three or more similar subjects. To solve it, we innovatively propose the overlap online detection and back-to-start sampling strategy, which first derives the most ambiguous subject and region, and then additionally imposes explicitly restriction loss $\mathcal{L}_{res}$, largely alleviating the overlapping issue when multiple similar subjects are generated.

\section{Conclusions}
\label{sec:conclusion}

In this paper, we focus on mitigating the similar subject generation problem for the most advanced MMDiT-based text-to-image model. Starting from the exploration of its generation mechanism, we identify three main ambiguities that exist during the generation process of MMDiT model, namely inter-block ambiguity, text encoder ambiguity and semantic ambiguity. To address these, block alignment loss, text encoder alignment loss, and overlap loss are tailored to repair the ambiguous latent by on-the-fly optimization in the early stage of denoising. In addition, we innovatively propose the overlap online detection and back-to-start sampling strategy, which imposes explicit restriction on the most ambiguous subject and region, thus further removing the stubborn semantic ambiguity during multiple similar subject generation. Extensive experiments and ablations demonstrate superiority of proposed methods. We hope that our exploration towards generation mechanism of MMDiT and proposed designs for mitigating generation problem will shed light on more future related generation and editing work.
\clearpage
{
    \small
    \bibliographystyle{ieeenat_fullname}
    \bibliography{main}

\begin{thebibliography}{43}
\providecommand{\natexlab}[1]{#1}
\providecommand{\url}[1]{\texttt{#1}}
\expandafter\ifx\csname urlstyle\endcsname\relax
  \providecommand{\doi}[1]{doi: #1}\else
  \providecommand{\doi}{doi: \begingroup \urlstyle{rm}\Url}\fi

\bibitem[Agarwal et~al.(2023)Agarwal, Karanam, Joseph, Saxena, Goswami, and Srinivasan]{agarwal2023star}
Aishwarya Agarwal, Srikrishna Karanam, KJ Joseph, Apoorv Saxena, Koustava Goswami, and Balaji~Vasan Srinivasan.
\newblock A-star: Test-time attention segregation and retention for text-to-image synthesis.
\newblock In \emph{Proceedings of the IEEE/CVF International Conference on Computer Vision}, pages 2283--2293, 2023.

\bibitem[Bao et~al.(2024)Bao, Li, Singh, Wang, and Hebert]{bao2024separate}
Zhipeng Bao, Yijun Li, Krishna~Kumar Singh, Yu-Xiong Wang, and Martial Hebert.
\newblock Separate-and-enhance: Compositional finetuning for text-to-image diffusion models.
\newblock In \emph{ACM SIGGRAPH 2024 Conference Papers}, pages 1--10, 2024.

\bibitem[Binyamin et~al.(2024)Binyamin, Tewel, Segev, Hirsch, Rassin, and Chechik]{binyamin2024make}
Lital Binyamin, Yoad Tewel, Hilit Segev, Eran Hirsch, Royi Rassin, and Gal Chechik.
\newblock Make it count: Text-to-image generation with an accurate number of objects.
\newblock \emph{arXiv preprint arXiv:2406.10210}, 2024.

\bibitem[Chefer et~al.(2023)Chefer, Alaluf, Vinker, Wolf, and Cohen-Or]{chefer2023attend}
Hila Chefer, Yuval Alaluf, Yael Vinker, Lior Wolf, and Daniel Cohen-Or.
\newblock Attend-and-excite: Attention-based semantic guidance for text-to-image diffusion models.
\newblock \emph{ACM Transactions on Graphics (TOG)}, 42\penalty0 (4):\penalty0 1--10, 2023.

\bibitem[Chen et~al.(2024)Chen, Laina, and Vedaldi]{chen2024training}
Minghao Chen, Iro Laina, and Andrea Vedaldi.
\newblock Training-free layout control with cross-attention guidance.
\newblock In \emph{Proceedings of the IEEE/CVF Winter Conference on Applications of Computer Vision}, pages 5343--5353, 2024.

\bibitem[Cherti et~al.(2023)Cherti, Beaumont, Wightman, Wortsman, Ilharco, Gordon, Schuhmann, Schmidt, and Jitsev]{cherti2023reproducible}
Mehdi Cherti, Romain Beaumont, Ross Wightman, Mitchell Wortsman, Gabriel Ilharco, Cade Gordon, Christoph Schuhmann, Ludwig Schmidt, and Jenia Jitsev.
\newblock Reproducible scaling laws for contrastive language-image learning.
\newblock In \emph{Proceedings of the IEEE/CVF Conference on Computer Vision and Pattern Recognition}, pages 2818--2829, 2023.

\bibitem[Dahary et~al.(2024)Dahary, Patashnik, Aberman, and Cohen-Or]{dahary2024yourself}
Omer Dahary, Or Patashnik, Kfir Aberman, and Daniel Cohen-Or.
\newblock Be yourself: Bounded attention for multi-subject text-to-image generation.
\newblock In \emph{European Conference on Computer Vision}. Springer, 2024.

\bibitem[Esser et~al.(2024)Esser, Kulal, Blattmann, Entezari, M{\"u}ller, Saini, Levi, Lorenz, Sauer, Boesel, et~al.]{esser2024scaling}
Patrick Esser, Sumith Kulal, Andreas Blattmann, Rahim Entezari, Jonas M{\"u}ller, Harry Saini, Yam Levi, Dominik Lorenz, Axel Sauer, Frederic Boesel, et~al.
\newblock Scaling rectified flow transformers for high-resolution image synthesis.
\newblock In \emph{Forty-first International Conference on Machine Learning}, 2024.

\bibitem[Feng et~al.()Feng, He, Fu, Jampani, Akula, Narayana, Basu, Wang, and Wang]{fengtraining}
Weixi Feng, Xuehai He, Tsu-Jui Fu, Varun Jampani, Arjun~Reddy Akula, Pradyumna Narayana, Sugato Basu, Xin~Eric Wang, and William~Yang Wang.
\newblock Training-free structured diffusion guidance for compositional text-to-image synthesis.
\newblock In \emph{The Eleventh International Conference on Learning Representations}.

\bibitem[Guo et~al.(2024)Guo, Liu, Cui, Li, Yang, and Huang]{guo2024initno}
Xiefan Guo, Jinlin Liu, Miaomiao Cui, Jiankai Li, Hongyu Yang, and Di Huang.
\newblock Initno: Boosting text-to-image diffusion models via initial noise optimization.
\newblock In \emph{Proceedings of the IEEE/CVF Conference on Computer Vision and Pattern Recognition}, pages 9380--9389, 2024.

\bibitem[Hao et~al.(2024)Hao, Chi, Dong, and Wei]{hao2024optimizing}
Yaru Hao, Zewen Chi, Li Dong, and Furu Wei.
\newblock Optimizing prompts for text-to-image generation.
\newblock \emph{Advances in Neural Information Processing Systems}, 36, 2024.

\bibitem[Heusel et~al.(2017)Heusel, Ramsauer, Unterthiner, Nessler, and Hochreiter]{Heusel2017GANsTB}
Martin Heusel, Hubert Ramsauer, Thomas Unterthiner, Bernhard Nessler, and Sepp Hochreiter.
\newblock Gans trained by a two time-scale update rule converge to a local nash equilibrium.
\newblock In \emph{NIPS}, 2017.

\bibitem[Ho and Salimans(2022)]{ho2022classifier}
Jonathan Ho and Tim Salimans.
\newblock Classifier-free diffusion guidance.
\newblock \emph{arXiv preprint arXiv:2207.12598}, 2022.

\bibitem[Ho et~al.(2020)Ho, Jain, and Abbeel]{ho2020denoising}
Jonathan Ho, Ajay Jain, and Pieter Abbeel.
\newblock Denoising diffusion probabilistic models.
\newblock \emph{Advances in neural information processing systems}, 33:\penalty0 6840--6851, 2020.

\bibitem[Kang et~al.(2023{\natexlab{a}})Kang, Zhu, Zhang, Park, Shechtman, Paris, and Park]{kang2023scaling}
Minguk Kang, Jun-Yan Zhu, Richard Zhang, Jaesik Park, Eli Shechtman, Sylvain Paris, and Taesung Park.
\newblock Scaling up gans for text-to-image synthesis.
\newblock In \emph{Proceedings of the IEEE/CVF Conference on Computer Vision and Pattern Recognition}, pages 10124--10134, 2023{\natexlab{a}}.

\bibitem[Kang et~al.(2023{\natexlab{b}})Kang, Galim, and Koo]{kang2023counting}
Wonjun Kang, Kevin Galim, and Hyung~Il Koo.
\newblock Counting guidance for high fidelity text-to-image synthesis.
\newblock \emph{arXiv preprint arXiv:2306.17567}, 2023{\natexlab{b}}.

\bibitem[Liu et~al.(2022)Liu, Li, Du, Torralba, and Tenenbaum]{liu2022compositional}
Nan Liu, Shuang Li, Yilun Du, Antonio Torralba, and Joshua~B Tenenbaum.
\newblock Compositional visual generation with composable diffusion models.
\newblock In \emph{European Conference on Computer Vision}, pages 423--439. Springer, 2022.

\bibitem[Liu et~al.(2024)Liu, Zeng, Ren, Li, Zhang, Yang, Jiang, Li, Yang, Su, Zhu, and Zhang]{liu2024grounding}
Shilong Liu, Zhaoyang Zeng, Tianhe Ren, Feng Li, Hao Zhang, Jie Yang, Qing Jiang, Chunyuan Li, Jianwei Yang, Hang Su, Jun Zhu, and Lei Zhang.
\newblock Grounding dino: Marrying dino with grounded pre-training for open-set object detection.
\newblock In \emph{European Conference on Computer Vision}. Springer, 2024.

\bibitem[Liu and Chilton(2022)]{liu2022design}
Vivian Liu and Lydia~B Chilton.
\newblock Design guidelines for prompt engineering text-to-image generative models.
\newblock In \emph{Proceedings of the 2022 CHI conference on human factors in computing systems}, pages 1--23, 2022.

\bibitem[Liu et~al.()Liu, Gong, et~al.]{liuflow}
Xingchao Liu, Chengyue Gong, et~al.
\newblock Flow straight and fast: Learning to generate and transfer data with rectified flow.
\newblock In \emph{The Eleventh International Conference on Learning Representations}.

\bibitem[Meral et~al.(2024)Meral, Simsar, Tombari, and Yanardag]{meral2024conform}
Tuna Han~Salih Meral, Enis Simsar, Federico Tombari, and Pinar Yanardag.
\newblock Conform: Contrast is all you need for high-fidelity text-to-image diffusion models.
\newblock In \emph{Proceedings of the IEEE/CVF Conference on Computer Vision and Pattern Recognition}, pages 9005--9014, 2024.

\bibitem[Nichol et~al.(2022)Nichol, Dhariwal, Ramesh, Shyam, Mishkin, Mcgrew, Sutskever, and Chen]{nichol2022glide}
Alexander~Quinn Nichol, Prafulla Dhariwal, Aditya Ramesh, Pranav Shyam, Pamela Mishkin, Bob Mcgrew, Ilya Sutskever, and Mark Chen.
\newblock Glide: Towards photorealistic image generation and editing with text-guided diffusion models.
\newblock In \emph{International Conference on Machine Learning}, pages 16784--16804. PMLR, 2022.

\bibitem[OpenAI(2024{\natexlab{a}})]{gpt4o_mini}
OpenAI.
\newblock Gpt-4o mini: Advancing cost-efficient intelligence.
\newblock \url{https://openai.com/index/gpt-4o-mini-advancing-cost-efficient-intelligence/}, 2024{\natexlab{a}}.
\newblock Accessed: 2024-11-03.

\bibitem[OpenAI(2024{\natexlab{b}})]{openai2024chatgpt}
OpenAI.
\newblock Chatgpt.
\newblock \url{https://chatgpt.com/}, 2024{\natexlab{b}}.
\newblock Version: GPT-4.

\bibitem[Peebles and Xie(2023)]{peebles2023scalable}
William Peebles and Saining Xie.
\newblock Scalable diffusion models with transformers.
\newblock In \emph{Proceedings of the IEEE/CVF International Conference on Computer Vision}, pages 4195--4205, 2023.

\bibitem[Podell et~al.(2023)Podell, English, Lacey, Blattmann, Dockhorn, M{\"u}ller, Penna, and Rombach]{podell2023sdxl}
Dustin Podell, Zion English, Kyle Lacey, Andreas Blattmann, Tim Dockhorn, Jonas M{\"u}ller, Joe Penna, and Robin Rombach.
\newblock Sdxl: Improving latent diffusion models for high-resolution image synthesis.
\newblock \emph{arXiv preprint arXiv:2307.01952}, 2023.

\bibitem[Radford et~al.(2021)Radford, Kim, Hallacy, Ramesh, Goh, Agarwal, Sastry, Askell, Mishkin, Clark, et~al.]{radford2021learning}
Alec Radford, Jong~Wook Kim, Chris Hallacy, Aditya Ramesh, Gabriel Goh, Sandhini Agarwal, Girish Sastry, Amanda Askell, Pamela Mishkin, Jack Clark, et~al.
\newblock Learning transferable visual models from natural language supervision.
\newblock In \emph{International conference on machine learning}, pages 8748--8763. PMLR, 2021.

\bibitem[Raffel et~al.(2020)Raffel, Shazeer, Roberts, Lee, Narang, Matena, Zhou, Li, and Liu]{raffel2020exploring}
Colin Raffel, Noam Shazeer, Adam Roberts, Katherine Lee, Sharan Narang, Michael Matena, Yanqi Zhou, Wei Li, and Peter~J Liu.
\newblock Exploring the limits of transfer learning with a unified text-to-text transformer.
\newblock \emph{Journal of machine learning research}, 21\penalty0 (140):\penalty0 1--67, 2020.

\bibitem[Ramesh et~al.()Ramesh, Dhariwal, Nichol, Chu, and Chen]{ramesh2022hierarchical}
Aditya Ramesh, Prafulla Dhariwal, Alex Nichol, Casey Chu, and Mark Chen.
\newblock Hierarchical text-conditional image generation with clip latents.

\bibitem[Rassin et~al.(2024)Rassin, Hirsch, Glickman, Ravfogel, Goldberg, and Chechik]{rassin2024linguistic}
Royi Rassin, Eran Hirsch, Daniel Glickman, Shauli Ravfogel, Yoav Goldberg, and Gal Chechik.
\newblock Linguistic binding in diffusion models: Enhancing attribute correspondence through attention map alignment.
\newblock \emph{Advances in Neural Information Processing Systems}, 36, 2024.

\bibitem[Rombach et~al.(2022)Rombach, Blattmann, Lorenz, Esser, and Ommer]{rombach2022high}
Robin Rombach, Andreas Blattmann, Dominik Lorenz, Patrick Esser, and Bj{\"o}rn Ommer.
\newblock High-resolution image synthesis with latent diffusion models.
\newblock In \emph{Proceedings of the IEEE/CVF conference on computer vision and pattern recognition}, pages 10684--10695, 2022.

\bibitem[Ronneberger et~al.(2015)Ronneberger, Fischer, and Brox]{ronneberger2015u}
Olaf Ronneberger, Philipp Fischer, and Thomas Brox.
\newblock U-net: Convolutional networks for biomedical image segmentation.
\newblock In \emph{Medical image computing and computer-assisted intervention--MICCAI 2015: 18th international conference, Munich, Germany, October 5-9, 2015, proceedings, part III 18}, pages 234--241. Springer, 2015.

\bibitem[Saharia et~al.(2022)Saharia, Chan, Saxena, Li, Whang, Denton, Ghasemipour, Gontijo~Lopes, Karagol~Ayan, Salimans, et~al.]{saharia2022photorealistic}
Chitwan Saharia, William Chan, Saurabh Saxena, Lala Li, Jay Whang, Emily~L Denton, Kamyar Ghasemipour, Raphael Gontijo~Lopes, Burcu Karagol~Ayan, Tim Salimans, et~al.
\newblock Photorealistic text-to-image diffusion models with deep language understanding.
\newblock \emph{Advances in neural information processing systems}, 35:\penalty0 36479--36494, 2022.

\bibitem[Song et~al.()Song, Meng, and Ermon]{songdenoising}
Jiaming Song, Chenlin Meng, and Stefano Ermon.
\newblock Denoising diffusion implicit models.
\newblock In \emph{International Conference on Learning Representations}.

\bibitem[Sueyoshi and Matsubara(2024)]{sueyoshi2024predicated}
Kota Sueyoshi and Takashi Matsubara.
\newblock Predicated diffusion: Predicate logic-based attention guidance for text-to-image diffusion models.
\newblock In \emph{Proceedings of the IEEE/CVF Conference on Computer Vision and Pattern Recognition}, pages 8651--8660, 2024.

\bibitem[Wang et~al.(2023)Wang, Montoya, Munechika, Yang, Hoover, and Chau]{wang2023diffusiondb}
Zijie~J Wang, Evan Montoya, David Munechika, Haoyang Yang, Benjamin Hoover, and Duen~Horng Chau.
\newblock Diffusiondb: A large-scale prompt gallery dataset for text-to-image generative models.
\newblock In \emph{The 61st Annual Meeting Of The Association For Computational Linguistics}, 2023.

\bibitem[Wei et~al.(2023)Wei, Chen, Zhou, Liao, Zhang, Hua, and Yu]{wei2023hairclipv2}
Tianyi Wei, Dongdong Chen, Wenbo Zhou, Jing Liao, Weiming Zhang, Gang Hua, and Nenghai Yu.
\newblock Hairclipv2: Unifying hair editing via proxy feature blending.
\newblock In \emph{Proceedings of the IEEE/CVF International Conference on Computer Vision}, pages 23589--23599, 2023.

\bibitem[Witteveen and Andrews(2022)]{witteveen2022investigating}
Sam Witteveen and Martin Andrews.
\newblock Investigating prompt engineering in diffusion models.
\newblock \emph{arXiv preprint arXiv:2211.15462}, 2022.

\bibitem[Xu et~al.(2018)Xu, Zhang, Huang, Zhang, Gan, Huang, and He]{xu2018attngan}
Tao Xu, Pengchuan Zhang, Qiuyuan Huang, Han Zhang, Zhe Gan, Xiaolei Huang, and Xiaodong He.
\newblock Attngan: Fine-grained text to image generation with attentional generative adversarial networks.
\newblock In \emph{Proceedings of the IEEE conference on computer vision and pattern recognition}, pages 1316--1324, 2018.

\bibitem[Yu et~al.(2022)Yu, Xu, Koh, Luong, Baid, Wang, Vasudevan, Ku, Yang, Ayan, et~al.]{yu2022scaling}
Jiahui Yu, Yuanzhong Xu, Jing~Yu Koh, Thang Luong, Gunjan Baid, Zirui Wang, Vijay Vasudevan, Alexander Ku, Yinfei Yang, Burcu~Karagol Ayan, et~al.
\newblock Scaling autoregressive models for content-rich text-to-image generation.
\newblock \emph{Transactions on Machine Learning Research}, 2022.

\bibitem[Zhang et~al.(2021)Zhang, Koh, Baldridge, Lee, and Yang]{zhang2021cross}
Han Zhang, Jing~Yu Koh, Jason Baldridge, Honglak Lee, and Yinfei Yang.
\newblock Cross-modal contrastive learning for text-to-image generation.
\newblock In \emph{Proceedings of the IEEE/CVF conference on computer vision and pattern recognition}, pages 833--842, 2021.

\bibitem[Zhang et~al.(2024{\natexlab{a}})Zhang, Tzun, Hern, Sim, and Kawaguchi]{zhang2024enhancing}
Yang Zhang, Teoh~Tze Tzun, Lim~Wei Hern, Tiviatis Sim, and Kenji Kawaguchi.
\newblock Enhancing semantic fidelity in text-to-image synthesis: Attention regulation in diffusion models.
\newblock In \emph{European Conference on Computer Vision}. Springer, 2024{\natexlab{a}}.

\bibitem[Zhang et~al.(2024{\natexlab{b}})Zhang, Yu, and Wu]{zhang2024object}
Yasi Zhang, Peiyu Yu, and Ying~Nian Wu.
\newblock Object-conditioned energy-based attention map alignment in text-to-image diffusion models.
\newblock In \emph{European Conference on Computer Vision}, pages 55--71. Springer, 2024{\natexlab{b}}.

\end{thebibliography}
}

\clearpage
\maketitlesupplementary

\section{Algorithm}
The pseudo-code for the entire denoising process using our method is provided in Algorithm~\ref{alg:our_method}.

\begin{algorithm}[h]
	\SetAlgoLined
	\KwIn{A text prompt $\mathcal{P}$, a pretrained MMDiT-based text-to-image model $\varepsilon_\theta$, an image decoder $\mathcal{D}$, total sampling steps $T$ (default $28$), denoising step $O$ (default $5$) when applying overlap online detection, iterations list $IterList$ (steps 1-2: 1 and steps 3-5: 15) for the first five denoising steps.}
	\KwOut{An image $x^{'}$ aligned with the text prompt $\mathcal{P}$ or rejecting sampling alerts.}
        \textbf{Initialize} $z_T \sim \mathcal{N}(0, 1)$, $z_{init} \leftarrow z_T$, BtS=\textit{False}.
        
	\For{$t=T$ to $T-O+1$}{
            \For{$n=1$ to $IterList[T-t]$}{
                $\_, A^{clip}, A^{t5} \leftarrow \varepsilon_\theta(z_t, t, \mathcal{P})$
                $z_{t}^{'} \leftarrow z_t - \alpha_{t}\nabla_{z_t}\mathcal{L}_{amb}(A^{clip}, A^{t5})$
            }
            $z_{t-1}, \_, \_ \leftarrow \varepsilon_\theta(z_{t}^{'}, t, \mathcal{P})$

        }
        \If{Overlap Online Detection == True}{
            BtS=\textit{True}, $z_T \leftarrow z_{init}$, 
            
            conflict region mask $M_{res}$ is derived.

    	\For{$t=T$ to $T-O+1$}{
                \For{$n=1$ to $IterList[T-t]$}{
                    $\_, A^{clip}, A^{t5} \leftarrow \varepsilon_\theta(z_t, t, \mathcal{P})$
                    $z_{t}^{'} \leftarrow z_t - \alpha_{t}\nabla_{z_t}\mathcal{L}(A^{clip}, A^{t5}, M_{res})$
                }
                $z_{t-1}, \_, \_ \leftarrow \varepsilon_\theta(z_{t}^{'}, t, \mathcal{P})$
    
            }
            \If{Overlap Online Detection == True}{
                \textbf{Return:} ``Bad seed, rejected sampling!"
            }
        }   
	\For{$t=T-O$ to $\frac{T}{2}+1$}{
            $\_, A^{clip}, A^{t5} \leftarrow \varepsilon_\theta(z_t, t, \mathcal{P})$

            \eIf{BtS}{
                $z_{t}^{'} \leftarrow z_t - \alpha_{t}\nabla_{z_t}\mathcal{L}(A^{clip}, A^{t5}, M_{res})$
            }
            {
                $z_{t}^{'} \leftarrow z_t - \alpha_{t}\nabla_{z_t}\mathcal{L}_{amb}(A^{clip}, A^{t5})$
            }
            $z_{t-1}, \_, \_ \leftarrow \varepsilon_\theta(z_{t}^{'}, t, \mathcal{P})$

        }

	\For{$t=\frac{T}{2}$ to $1$}{
            $z_{t-1}, \_, \_ \leftarrow \varepsilon_\theta(z_{t}, t, \mathcal{P})$

        }
        $x^{'} \leftarrow \mathcal{D}(z_0) $
        
        \textbf{Return:} $x^{'}$

	\caption{Entire Denoising Process Using Our Method}
	\label{alg:our_method}
\end{algorithm}

\section{Implementation Details of Metrics}
\subsection{SR (G DINO)}
SR (G DINO) represents the Success Rate (whether or not the subjects all appear in the results) calculated using the state-of-the-art open-set object detection method Grounding DINO~\cite{liu2024grounding}. Specifically, for an image synthesized from the given text prompt (e.g., ``a chicken and a duck and a goose"), we first parse the prompt to obtain the text hint (``chicken . duck. goose .") of objects to be detected required by Grounding DINO. Subsequently, this hint is fed into Grounding DINO along with the synthesized image to obtain the confidence value (a float number between $0$ and $1$) for the presence of each subject. If the confidence value of any subject is 0, the success rate of the image is 0; if the confidence values of all the subjects are greater than a threshold, the success rate of the image is 1; and for the remaining cases, the success rate of the image is the average of the confidence values of all the subjects. For two, three, and four similar subject generation settings, the thresholds are set to $0.5$, $0.4$, and $0.4$, respectively. For each text prompt, each method synthesizes $100$ results. We obtain their respective average success rates. If the average success rate of all methods is lower than $0.2$, we consider that there is an object or objects in the text prompt that are difficult to be detected by Grounding DINO, and therefore this text prompt will be excluded from the final success rate statistics.
\subsection{SR (GPT 4o)}
SR (GPT 4o) represents the Success Rate (whether or not the subjects all appear in the results) calculated using the most powerful and efficient intelligent model GPT 4o-mini~\cite{gpt4o_mini}. For a $256 \times 256$ image, the detection price of calling the GPT 4o-mini API is about $\$0.0013$. We resize the synthesized results to $256 \times 256$ and then feed them to GPT 4o-mini for detection. Again taking ``a chicken and a duck and a goose" as an example, the text input we constructed for GPT 4o-mini is: ``Please analyze the image data and determine if it contains all of the following subjects: chicken, duck, goose. Respond with `yes' if all subjects are present, and `no' if any are missing". Subsequently, `yes' is parsed as $1$ and `no' is parsed as $0$ to be the success rate of that image. As stated in the main text, to save cost, for each text prompt we randomly sampled $50$ groups of results from the original $100$ groups of different methods to call the GPT 4o-mini detection, which does not affect the statistical accuracy.

\section{Ablation Analysis}
\subsection{Ablation on the Step Range and Number of Iterations for Multiple Optimization}

\begin{figure}
    \centering
    \begin{subfigure}[t]{\columnwidth}
        \centering
        \includegraphics[width=\columnwidth]{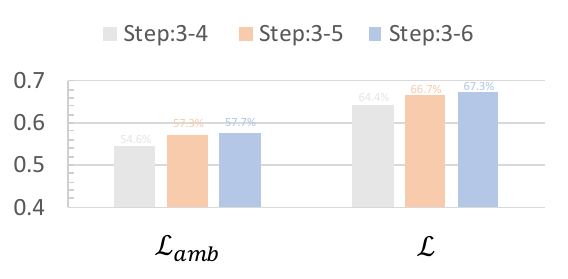} 
        \caption{Ablation of the step range. All iterations are set to $15$.}
        \label{fig:abltion_step}
    \end{subfigure}
    \begin{subfigure}[t]{\columnwidth}
        \centering
        \includegraphics[width=\columnwidth]{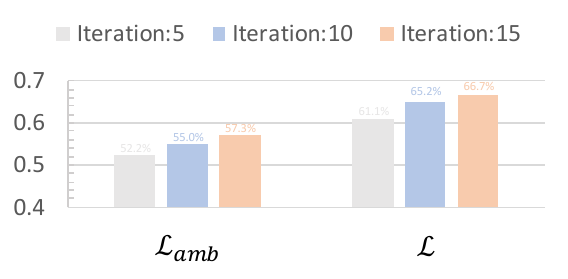}
        \caption{Ablation of iteration numbers. The step range is 3-5 for all.}
        \label{fig:ablation_iter}
    \end{subfigure}
    \caption{Ablation on the step range and number of iterations for multiple optimization. We provide the SR (G DINO) results of whether or not to apply the overlap online detection and back-to-start sampling strategy, \ie, $\mathcal{L}_{amb}$, $\mathcal{L}=\mathcal{L}_{amb}+\mathcal{L}_{res}$, respectively.}
    \label{fig:ablation_step_iter}
\end{figure}

We choose to perform multiple iterations of optimization in the early stages of sampling to enhance the effect of the repair. Quantitative ablations regarding the step range and the number of iterations are provided in Figure~\ref{fig:abltion_step} and~\ref{fig:ablation_iter}, respectively. Consistent with the main text, this ablation analysis is performed using SR (G DINO) metric on the most representative three similar subjects dataset. We provide the results of whether or not to apply the overlap online detection and back-to-start sampling strategy, \ie, $\mathcal{L}_{amb}$, $\mathcal{L}=\mathcal{L}_{amb}+\mathcal{L}_{res}$, respectively. The results of the two settings demonstrate a consistent trend. Balancing performance and runtime, we choose the step range as 3-5 and the number of iterations as 15 to be our default settings.

\section{Experiments}

\subsection{More Qualitative Results}
In Figures~\ref{fig:quail_comparsion_1} and~\ref{fig:quail_comparsion_2}, we provide more qualitative comparison results of our method with other state of the art methods on the constructed challenging similar-subject prompt dataset.

\begin{figure*}[t]
	\centering
	\includegraphics[width=\textwidth]{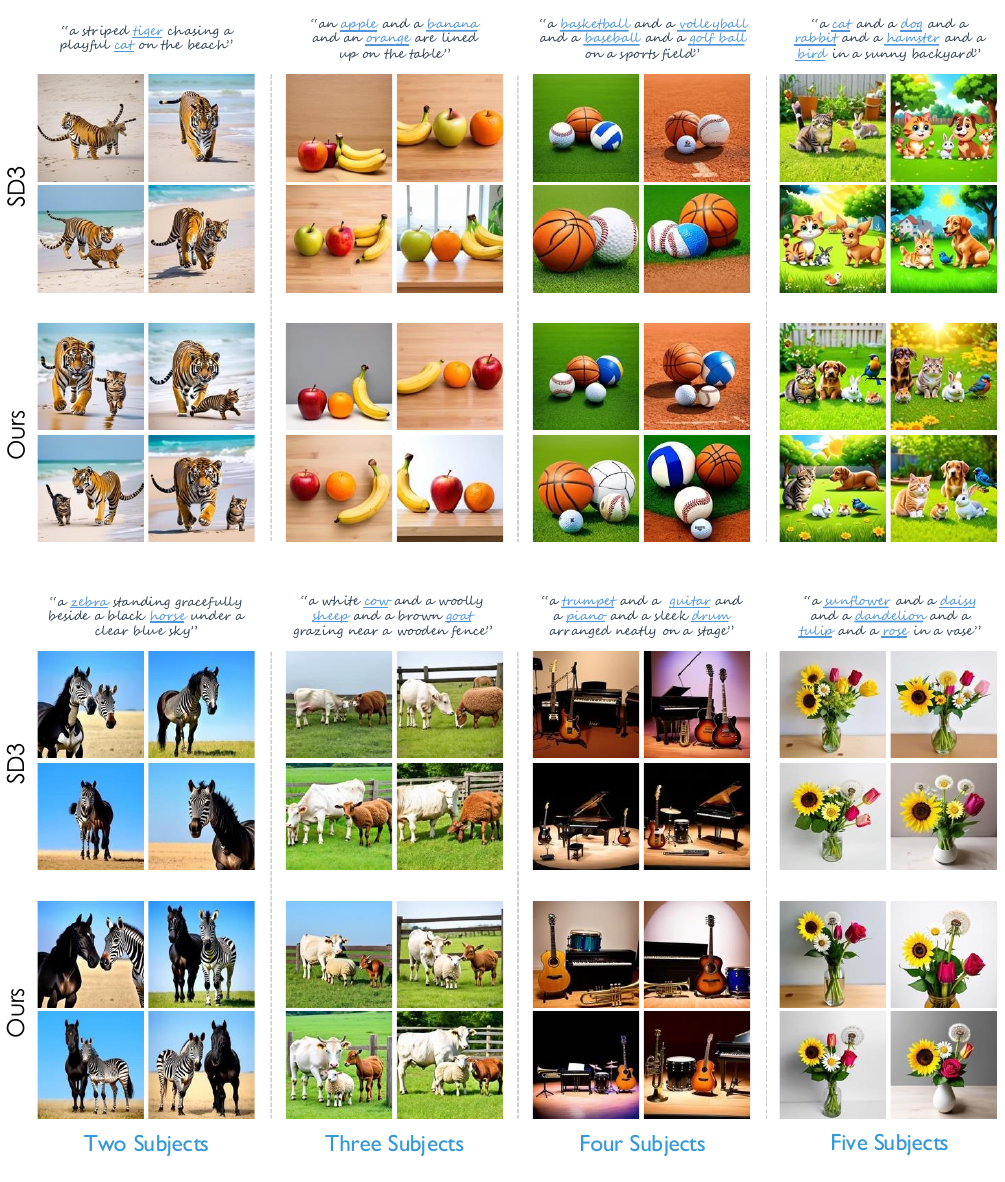}
	\caption{Illustration of applying our method on various complex prompts. Here, we provide visual results for complex prompts containing two, three, four, and five similar subjects.} 
	\label{fig:supp_teaser}
\end{figure*}

In addition, we provide the results of our approach on complex prompts in Figure~\ref{fig:supp_teaser}. As stated in the main text, the simple construction of similar-subject prompts in the dataset is just for the convenience of quantitative evaluation and illustration. Figure~\ref{fig:supp_teaser} demonstrates that our method can be applied to a variety of complex prompts.
\subsection{Time Consumption for Optimization}
\begin{table}[h]
    \centering
    \begin{tabular}{l|cccccc}
        \hline
        \small{Methods} & \small{SD3} &\small{~\cite{chefer2023attend}}&\small{~\cite{zhang2024object}}&\small{~\cite{meral2024conform}}& \small{$\mathcal{L}_{amb}$} & \small{$\mathcal{L}$} \\
        \hline
        \small{Time(s)} & \small{20.4} & \small{71.2} & \small{28.6} & \small{68.8} & \small{68.7} & \small{118.6} \\
        \hline
    \end{tabular}
    \caption{Report on the average time (in seconds) required to sample (on the NVIDIA A6000 GPU) an image by different methods.}
    \label{tab:time_quan_comparsion}
\end{table}
We provide in Table~\ref{tab:time_quan_comparsion} the average time required for our method to sample (on the NVIDIA A6000 GPU) an image with ($\mathcal{L}=\mathcal{L}_{amb}+\mathcal{L}_{res}$) or without ($\mathcal{L}_{amb}$) applying the back-to-start sampling strategy, as well as for other methods. It is worth mentioning that even without applying the back-to-start sampling strategy, our $\mathcal{L}_{amb}$ achieves the best repair success rate while our time overhead is only second to EBAMA~\cite{zhang2024object}. After applying the back-to-start sampling strategy, the success rate of our method is further improved (substantially ahead of other methods), albeit with additional time overhead. In addition, our back-to-start sampling strategy is only applied when it is triggered (\ie, the overlap is detected by the overlap online detection). For two, three, and four similar-subject datasets, the probabilities of being triggered are $0.4\%$, $29.3\%$, and $42.1\%$, respectively, which means that our full algorithm will take less than $118.6$ seconds to sample an image. For example, for two similar subjects, our sampling time is about $68.7$ seconds.

\subsection{Results on SD3.5}

\begin{table}[h]
    \centering
    \begin{subtable}[h]{\columnwidth}
        \centering
        \small
        \setlength{\tabcolsep}{9pt}{
        \begin{tabular}{l|ll}
            \hline
            Methods & SR (G DINO) $\uparrow$ & FID $\downarrow$ \\ \hline
            SD3.5~\cite{esser2024scaling}  & 52.0\% & - \\
            A\&E~\cite{chefer2023attend}  & 55.1\% (\textcolor{blue}{+3.1\%}) & 7.19 \\
            EBAMA~\cite{zhang2024object}  & 56.2\% (\textcolor{blue}{+4.2\%}) & 19.34 \\
            CONFORM~\cite{meral2024conform}  & 62.4\% (\textcolor{blue}{+10.4\%}) & 29.47 \\
            Ours-(w/o RS)  & \textbf{65.6\%} (\textcolor{blue}{+13.6\%}) & 18.39 \\
            Ours-(w/ RS)  & \textbf{65.6\%} (\textcolor{blue}{+13.6\%}) & 18.39 \\ \hline
        \end{tabular}
            }
        \caption{Two Similar Subjects}
        \label{tab:sd3point5_quan_subtable1}
    \end{subtable}
    \hspace{1em}

    \begin{subtable}[h]{\columnwidth}
        \centering
        \small
        \setlength{\tabcolsep}{9pt}{
        \begin{tabular}{l|ll}
            \hline
            Methods & SR (G DINO) $\uparrow$ & FID $\downarrow$ \\ \hline
            SD3.5~\cite{esser2024scaling}  & 34.0\% & - \\
            A\&E~\cite{chefer2023attend}  & 34.0\% (\textcolor{blue}{+0.0\%}) & 5.88 \\
            EBAMA~\cite{zhang2024object}  & 44.9\% (\textcolor{blue}{+10.9\%}) & 20.26 \\
            CONFORM~\cite{meral2024conform}  & 53.7\% (\textcolor{blue}{+19.7\%}) & 34.59 \\
            Ours-(w/o RS)  & \textbf{64.8\%} (\textcolor{blue}{+30.8\%}) & 19.44 \\
            Ours-(w/ RS)  & \textbf{69.4\%} (\textcolor{blue}{+35.4\%}) & 20.98 \\ \hline
        \end{tabular}
            }
        \caption{Three Similar Subjects}
        \label{tab:sd3point5_quan_subtable2}
    \end{subtable}
    \hspace{1em}

    \begin{subtable}[h]{\columnwidth}
        \centering
        \small
        \setlength{\tabcolsep}{9pt}{
        \begin{tabular}{l|ll}
            \hline
            Methods & SR (G DINO) $\uparrow$ & FID $\downarrow$ \\ \hline
            SD3.5~\cite{esser2024scaling}  & 26.2\% & - \\
            A\&E~\cite{chefer2023attend}  & 28.1\% (\textcolor{blue}{+1.9\%}) & 4.23 \\
            EBAMA~\cite{zhang2024object}  & 35.5\% (\textcolor{blue}{+9.3\%}) & 22.57 \\
            CONFORM~\cite{meral2024conform}  & 36.1\% (\textcolor{blue}{+9.9\%}) & 35.68 \\
            Ours-(w/o RS)  & \textbf{55.2\%} (\textcolor{blue}{+29.0\%}) & 27.79 \\
            Ours-(w/ RS)  & \textbf{65.9\%} (\textcolor{blue}{+39.7\%}) & 28.74 \\ \hline
        \end{tabular}
            }
        \caption{Four Similar Subjects}
        \label{tab:sd3point5_quan_subtable3}
    \end{subtable}

    \caption{Quantitative comparison with A\&E~\cite{chefer2023attend}, EBAMA~\cite{zhang2024object}, CONFORM~\cite{meral2024conform} on SD3.5~\cite{esser2024scaling}. SR (G DINO) represents the Success Rate (whether or not the subjects all appear in the results) calculated using Grounding DINO~\cite{liu2024grounding}. Blue numbers indicate the percent gain of different methods against SD3.5. FID~\cite{Heusel2017GANsTB} is calculated between the different methods and the sampling results of SD3.5, representing the ability of different methods to maintain the original generation quality of SD3.5. Ours-(w/ RS) and Ours-(w/o RS) represent whether or not our method is applied with the Reject Sampling strategy, respectively.}
    \label{tab:sd3point5_quan}
\end{table}

\begin{figure*}[t]
	\begin{center}
		\setlength{\tabcolsep}{0.5pt}
		\begin{tabular}{m{0.3cm}<{\centering}m{1.68cm}<{\centering}m{1.68cm}<{\centering}m{1.68cm}<{\centering}m{1.68cm}<{\centering}m{1.68cm}<{\centering}m{1.68cm}<{\centering}m{1.68cm}<{\centering}m{1.68cm}<{\centering}m{1.68cm}<{\centering}m{1.68cm}<{\centering}}
			 & \multicolumn{2}{c}{\scriptsize{\textit{``a \underline{snake} and a \underline{lizard}"}}} & \multicolumn{2}{c}{\scriptsize{\makecell{\scriptsize{\textit{``a \underline{watermelon} and a \underline{muskmelon}}} \\ \scriptsize{\textit{and a \underline{coconut}"}}}}} & \multicolumn{2}{c}{\scriptsize{\makecell{\scriptsize{\textit{``a \underline{cat} and a \underline{tiger}}} \\ \scriptsize{\textit{and a \underline{rabbit}"}}}}} & \multicolumn{2}{c}{\scriptsize{\makecell{\scriptsize{\textit{``a \underline{bike} and a \underline{scooter} and}} \\ \scriptsize{\textit{a \underline{skateboard} and a \underline{car}"}}}}} & \multicolumn{2}{c}{\scriptsize{\makecell{\scriptsize{\textit{``a \underline{lion} and a \underline{tiger} and}} \\ \scriptsize{\textit{a \underline{bear} and a \underline{wolf}"}}}}}
			\\
                \noalign{\vskip 3pt}
   
			\multirow{2}{*}{\raisebox{-0.85cm}{\rotatebox[origin=c]{90}{\footnotesize{{SD3.5}}}}}
			&\includegraphics[width=1.65cm]{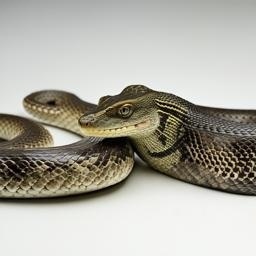}
			&\includegraphics[width=1.65cm]{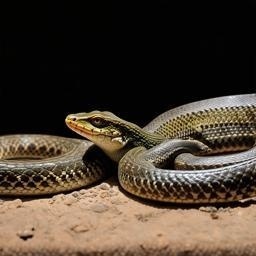}
			&\includegraphics[width=1.65cm]{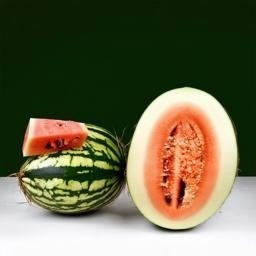}
			&\includegraphics[width=1.65cm]{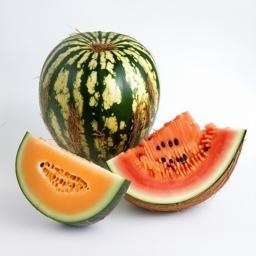}
			&\includegraphics[width=1.65cm]{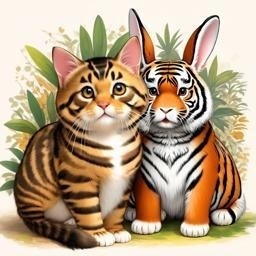}
			&\includegraphics[width=1.65cm]{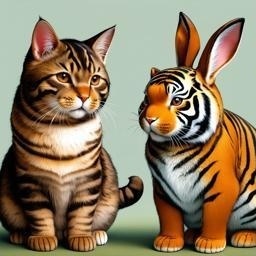}
			&\includegraphics[width=1.65cm]{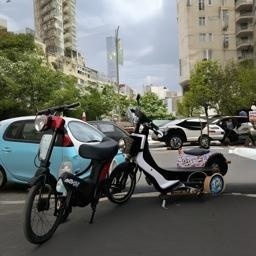}
			&\includegraphics[width=1.65cm]{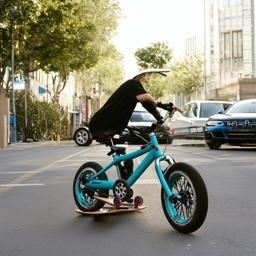}
			&\includegraphics[width=1.65cm]{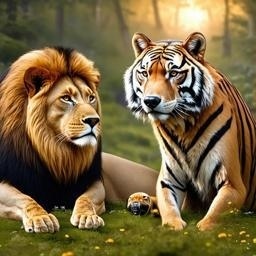}
			&\includegraphics[width=1.65cm]{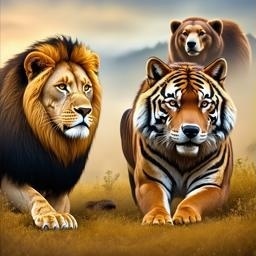}
			\\

			&\includegraphics[width=1.65cm]{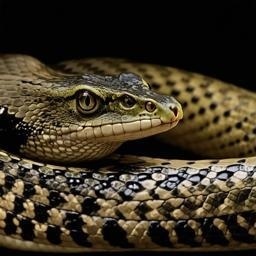}
			&\includegraphics[width=1.65cm]{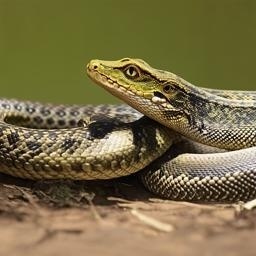}
			&\includegraphics[width=1.65cm]{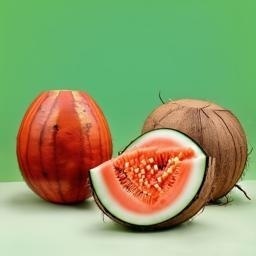}
			&\includegraphics[width=1.65cm]{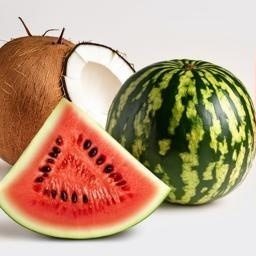}
			&\includegraphics[width=1.65cm]{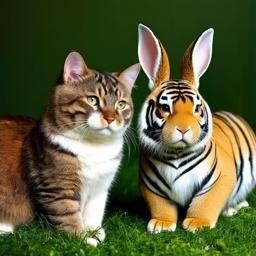}
			&\includegraphics[width=1.65cm]{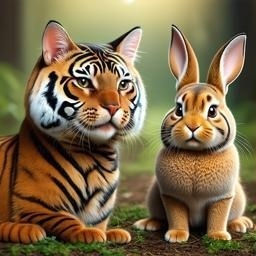}
			&\includegraphics[width=1.65cm]{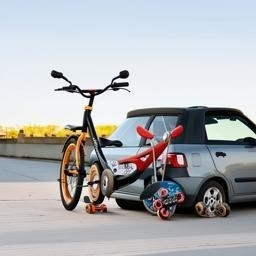}
			&\includegraphics[width=1.65cm]{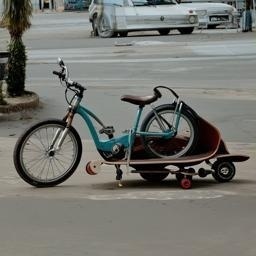}
			&\includegraphics[width=1.65cm]{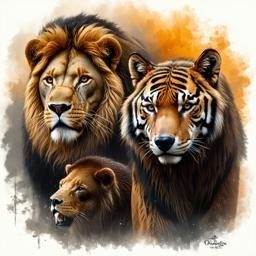}
			&\includegraphics[width=1.65cm]{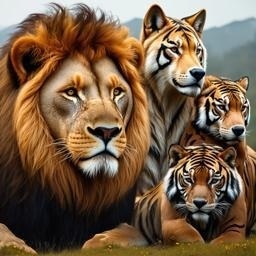}
			\\

                \\

			\multirow{2}{*}{\raisebox{-0.85cm}{\rotatebox[origin=c]{90}{\footnotesize{{A\&E}}}}}
			&\includegraphics[width=1.65cm]{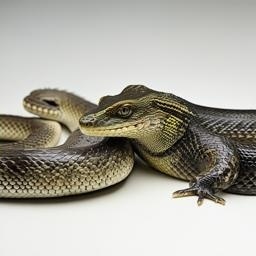}
			&\includegraphics[width=1.65cm]{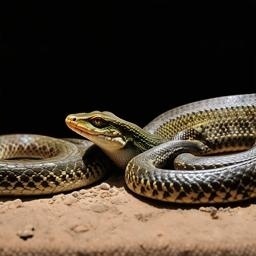}
			&\includegraphics[width=1.65cm]{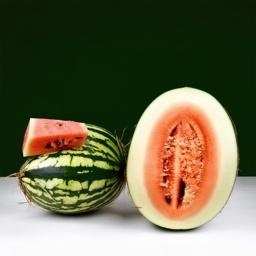}
			&\includegraphics[width=1.65cm]{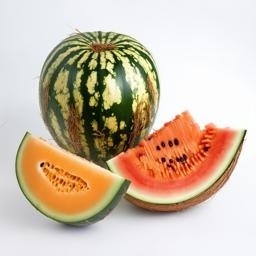}
			&\includegraphics[width=1.65cm]{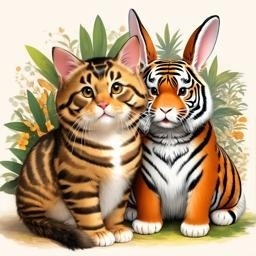}
			&\includegraphics[width=1.65cm]{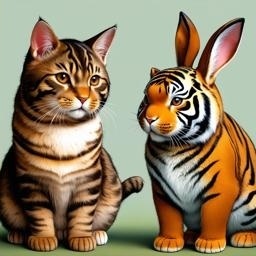}
			&\includegraphics[width=1.65cm]{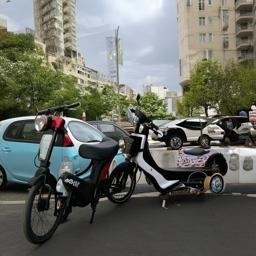}
			&\includegraphics[width=1.65cm]{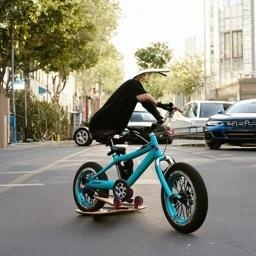}
			&\includegraphics[width=1.65cm]{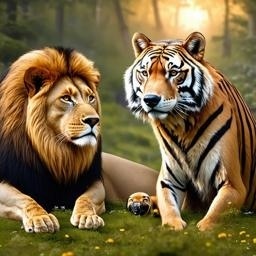}
			&\includegraphics[width=1.65cm]{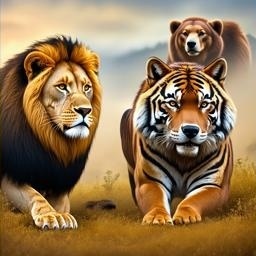}
			\\

			&\includegraphics[width=1.65cm]{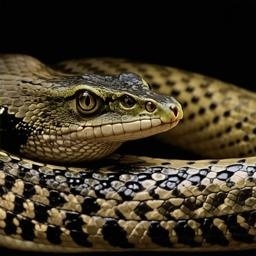}
			&\includegraphics[width=1.65cm]{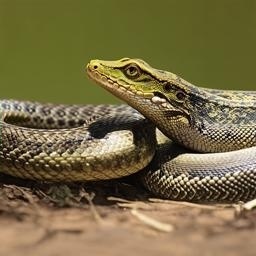}
			&\includegraphics[width=1.65cm]{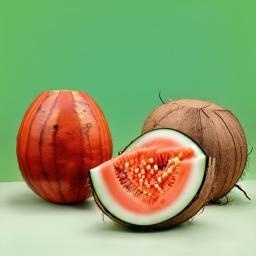}
			&\includegraphics[width=1.65cm]{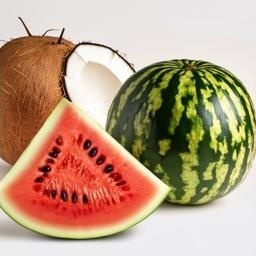}
			&\includegraphics[width=1.65cm]{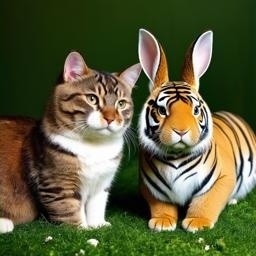}
			&\includegraphics[width=1.65cm]{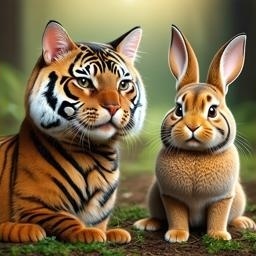}
			&\includegraphics[width=1.65cm]{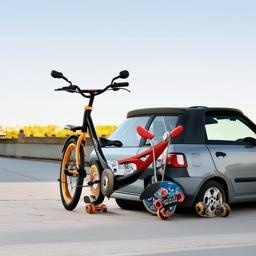}
			&\includegraphics[width=1.65cm]{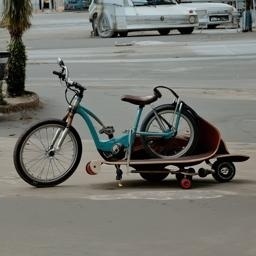}
			&\includegraphics[width=1.65cm]{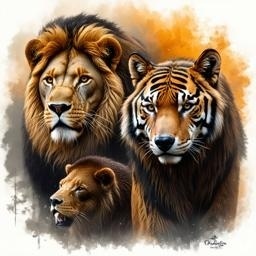}
			&\includegraphics[width=1.65cm]{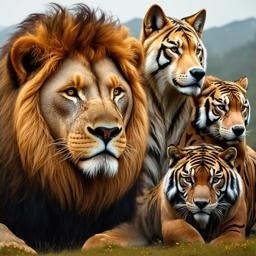}
			\\

                \\

			\multirow{2}{*}{\raisebox{-0.85cm}{\rotatebox[origin=c]{90}{\footnotesize{{EBAMA}}}}}
			&\includegraphics[width=1.65cm]{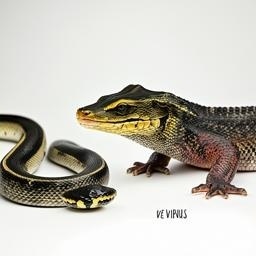}
			&\includegraphics[width=1.65cm]{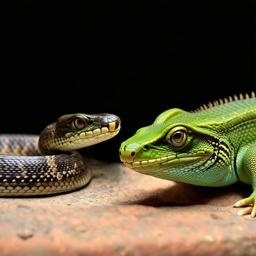}
			&\includegraphics[width=1.65cm]{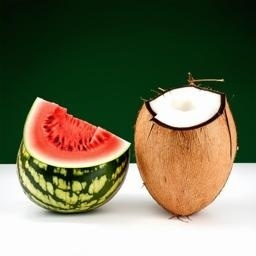}
			&\includegraphics[width=1.65cm]{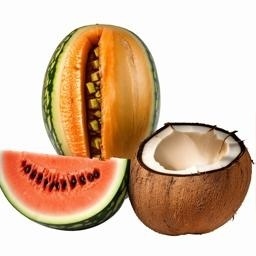}
			&\includegraphics[width=1.65cm]{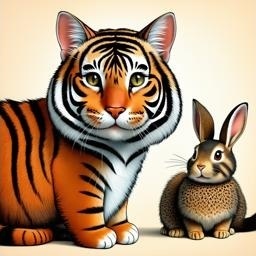}
			&\includegraphics[width=1.65cm]{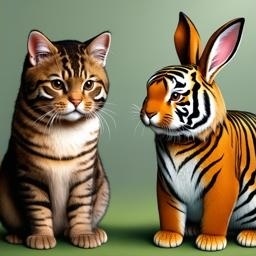}
			&\includegraphics[width=1.65cm]{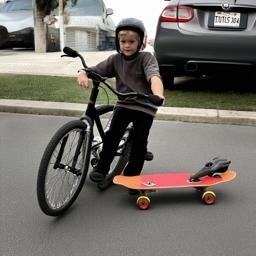}
			&\includegraphics[width=1.65cm]{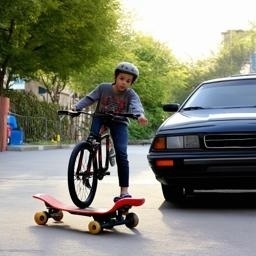}
			&\includegraphics[width=1.65cm]{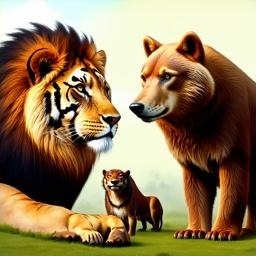}
			&\includegraphics[width=1.65cm]{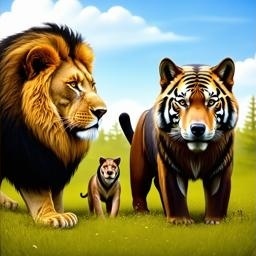}
			\\
		
			&\includegraphics[width=1.65cm]{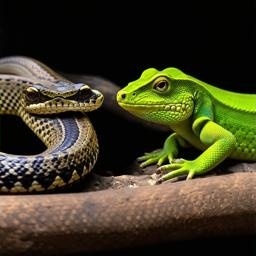}
			&\includegraphics[width=1.65cm]{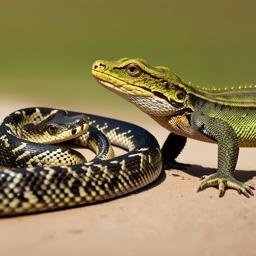}
			&\includegraphics[width=1.65cm]{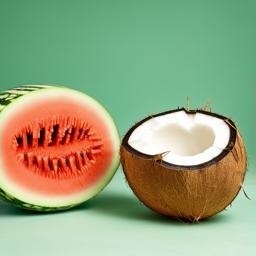}
			&\includegraphics[width=1.65cm]{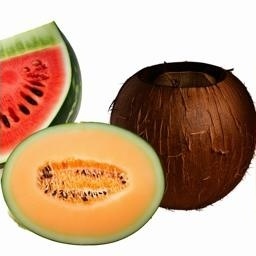}
			&\includegraphics[width=1.65cm]{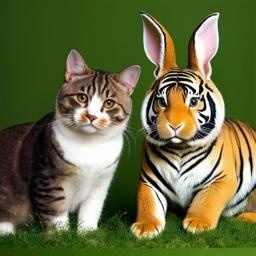}
			&\includegraphics[width=1.65cm]{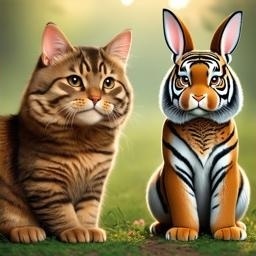}
			&\includegraphics[width=1.65cm]{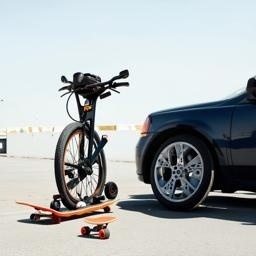}
			&\includegraphics[width=1.65cm]{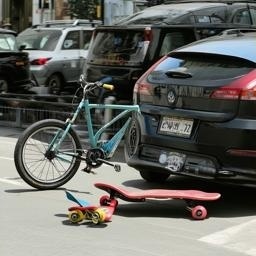}
			&\includegraphics[width=1.65cm]{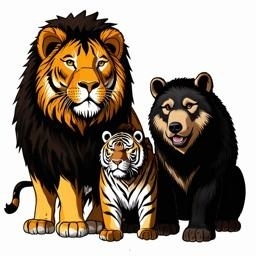}
			&\includegraphics[width=1.65cm]{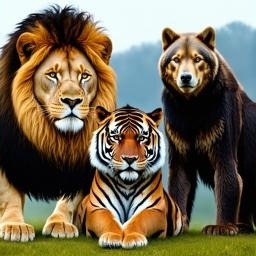}
			\\

                \\
                
			\multirow{2}{*}{\raisebox{-0.85cm}{\rotatebox[origin=c]{90}{\footnotesize{{CONFORM}}}}}
			&\includegraphics[width=1.65cm]{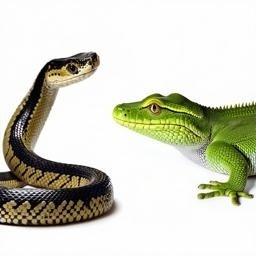}
			&\includegraphics[width=1.65cm]{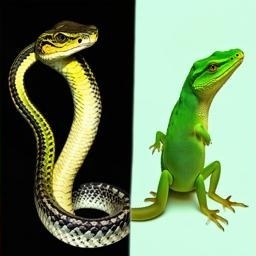}
			&\includegraphics[width=1.65cm]{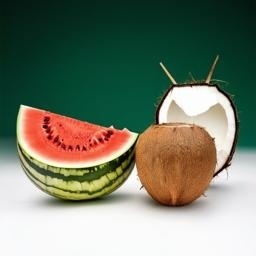}
			&\includegraphics[width=1.65cm]{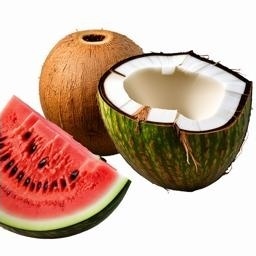}
			&\includegraphics[width=1.65cm]{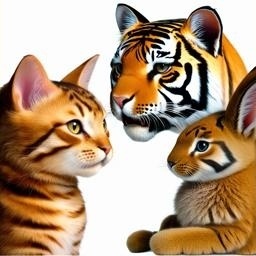}
			&\includegraphics[width=1.65cm]{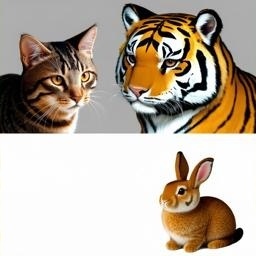}
			&\includegraphics[width=1.65cm]{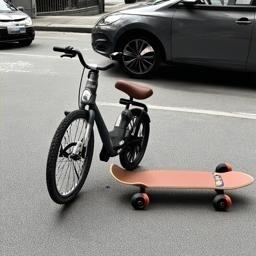}
			&\includegraphics[width=1.65cm]{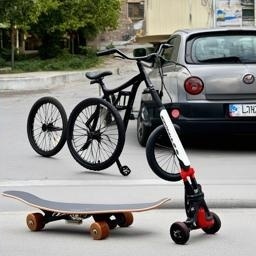}
			&\includegraphics[width=1.65cm]{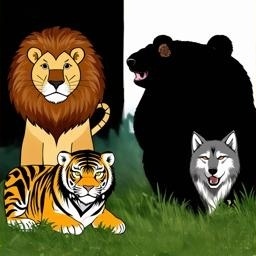}
			&\includegraphics[width=1.65cm]{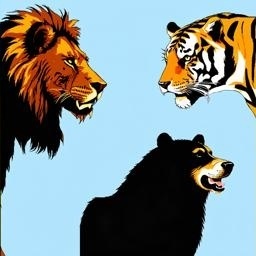}
			\\

			&\includegraphics[width=1.65cm]{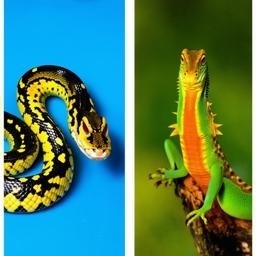}
			&\includegraphics[width=1.65cm]{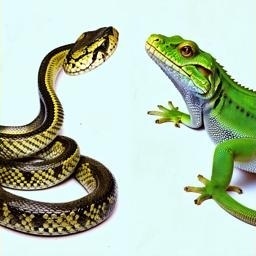}
			&\includegraphics[width=1.65cm]{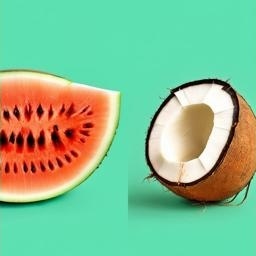}
			&\includegraphics[width=1.65cm]{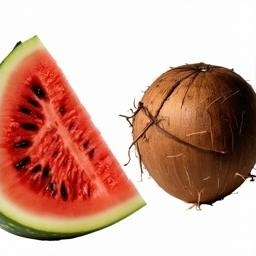}
			&\includegraphics[width=1.65cm]{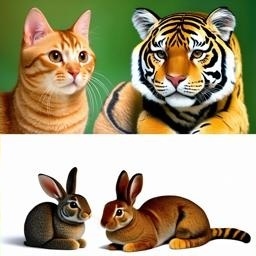}
			&\includegraphics[width=1.65cm]{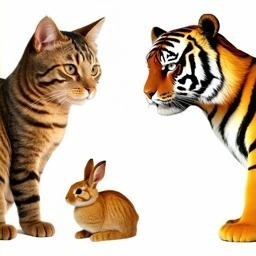}
			&\includegraphics[width=1.65cm]{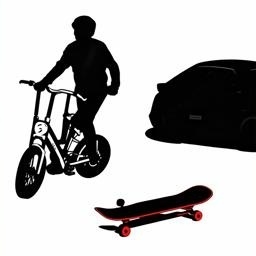}
			&\includegraphics[width=1.65cm]{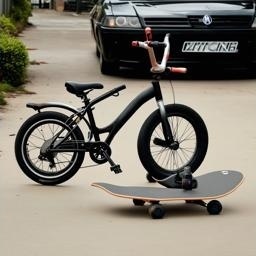}
			&\includegraphics[width=1.65cm]{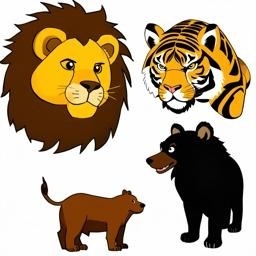}
			&\includegraphics[width=1.65cm]{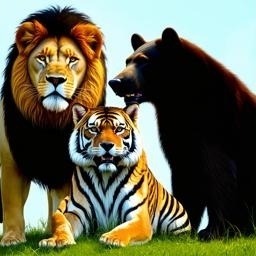}
			\\

                \\

			\multirow{2}{*}{\raisebox{-0.85cm}{\rotatebox[origin=c]{90}{\footnotesize{{Ours}}}}}
			&\includegraphics[width=1.65cm]{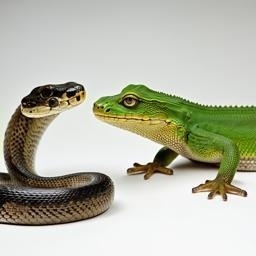}
			&\includegraphics[width=1.65cm]{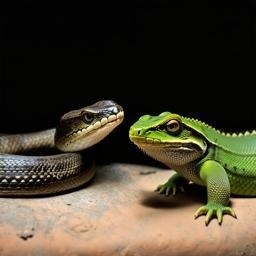}
			&\includegraphics[width=1.65cm]{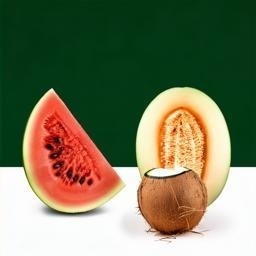}
			&\includegraphics[width=1.65cm]{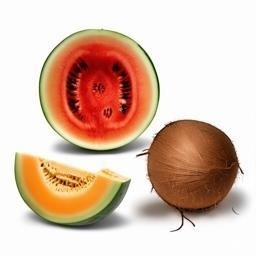}
			&\includegraphics[width=1.65cm]{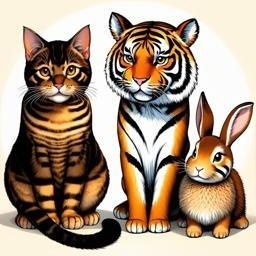}
			&\includegraphics[width=1.65cm]{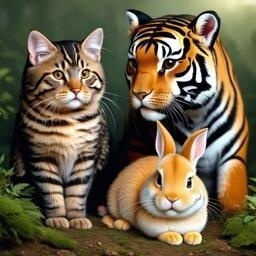}
			&\includegraphics[width=1.65cm]{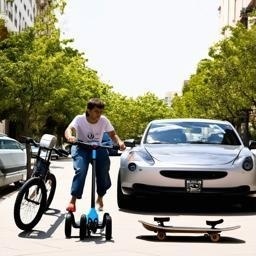}
			&\includegraphics[width=1.65cm]{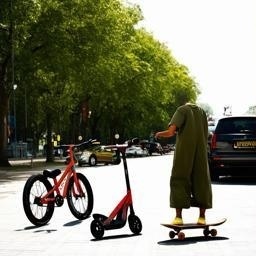}
			&\includegraphics[width=1.65cm]{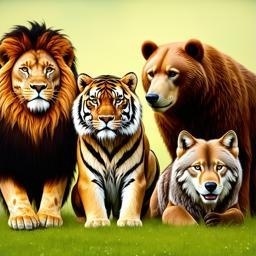}
			&\includegraphics[width=1.65cm]{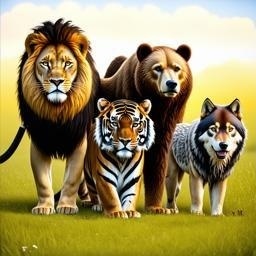}
			\\

			&\includegraphics[width=1.65cm]{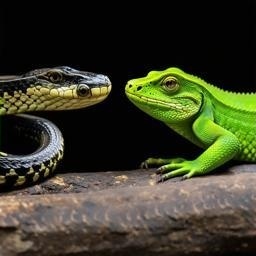}
			&\includegraphics[width=1.65cm]{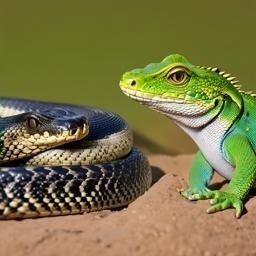}
			&\includegraphics[width=1.65cm]{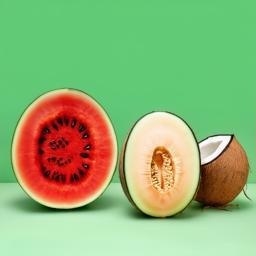}
			&\includegraphics[width=1.65cm]{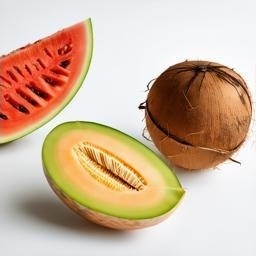}
			&\includegraphics[width=1.65cm]{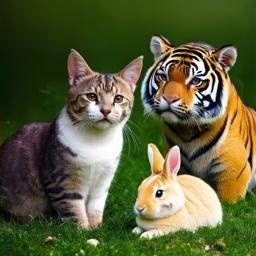}
			&\includegraphics[width=1.65cm]{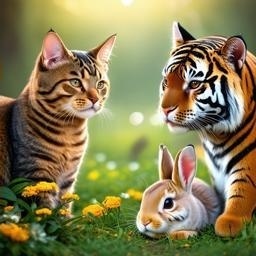}
			&\includegraphics[width=1.65cm]{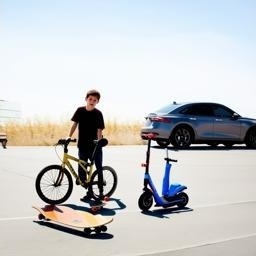}
			&\includegraphics[width=1.65cm]{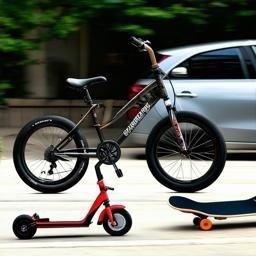}
			&\includegraphics[width=1.65cm]{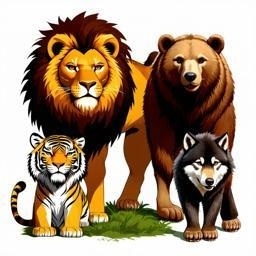}
			&\includegraphics[width=1.65cm]{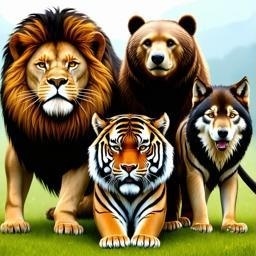}
			\\
            
		\end{tabular}
	\end{center}
	\caption{Qualitative comparison with A\&E~\cite{chefer2023attend}, EBAMA~\cite{zhang2024object}, CONFORM~\cite{meral2024conform} on SD3.5~\cite{esser2024scaling} using prompts containing two, three, and four similar subjects. Our approach mitigates the subject neglect or mixing problems present in SD3.5 while maintaining the superior generation quality.}
	\label{fig:sd3point5_quail_comparsion}
\end{figure*}

Recently, the MMDiT-based text-to-image model Stable Diffusion 3.5~\cite{esser2024scaling} has been released. We further demonstrate the generality of our method on SD3.5-Medium. Compared to SD3, the main improvement in the network structure of SD3.5 is that it introduces image (\ie, latent hidden representation) self-attention modules in the first $13$ layers of the transformer, which enhances multi-resolution generation and overall image coherence. We compare our method qualitatively and quantitatively with other state-of-the-art methods on SD3.5 using the constructed similar-subject prompt dataset. As shown in Table~\ref{tab:sd3point5_quan} and Figure~\ref{fig:sd3point5_quail_comparsion}, the subject neglect or mixing issue when generating similar subjects is not alleviated on SD3.5, while our approach still largely mitigates the issue while ensuring superior generation quality.

\begin{figure*}[t]
	\begin{center}
		\setlength{\tabcolsep}{0.5pt}
		\begin{tabular}{m{0.3cm}<{\centering}m{1.68cm}<{\centering}m{1.68cm}<{\centering}m{1.68cm}<{\centering}m{1.68cm}<{\centering}m{1.68cm}<{\centering}m{1.68cm}<{\centering}m{1.68cm}<{\centering}m{1.68cm}<{\centering}m{1.68cm}<{\centering}m{1.68cm}<{\centering}}
			 & \multicolumn{2}{c}{\scriptsize{\textit{``a \underline{whale} and a \underline{manatee}"}}} & \multicolumn{2}{c}{\scriptsize{\makecell{\scriptsize{\textit{``a \underline{tomato} and an \underline{apple}}} \\ \scriptsize{\textit{and a \underline{peach}"}}}}} & \multicolumn{2}{c}{\scriptsize{\makecell{\scriptsize{\textit{``a \underline{deer} and a \underline{llama}}} \\ \scriptsize{\textit{and a \underline{camel}"}}}}} & \multicolumn{2}{c}{\scriptsize{\makecell{\scriptsize{\textit{``a \underline{basketball} and a \underline{soccerball} and}} \\ \scriptsize{\textit{a \underline{baseball} and a \underline{tennisball}"}}}}} & \multicolumn{2}{c}{\scriptsize{\makecell{\scriptsize{\textit{``a \underline{chicken} and a \underline{duck} and}} \\ \scriptsize{\textit{a \underline{turtle} and a \underline{goose}"}}}}}
			\\
                \noalign{\vskip 3pt}
   
			\multirow{2}{*}{\raisebox{-0.85cm}{\rotatebox[origin=c]{90}{\footnotesize{{SD3}}}}}
			&\includegraphics[width=1.65cm]{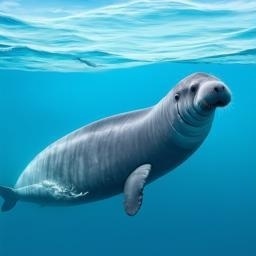}
			&\includegraphics[width=1.65cm]{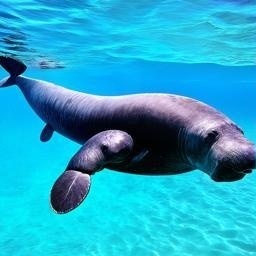}
			&\includegraphics[width=1.65cm]{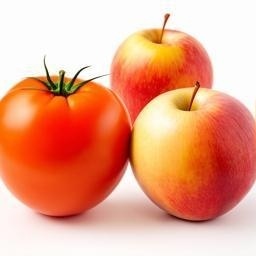}
			&\includegraphics[width=1.65cm]{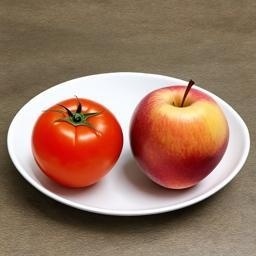}
			&\includegraphics[width=1.65cm]{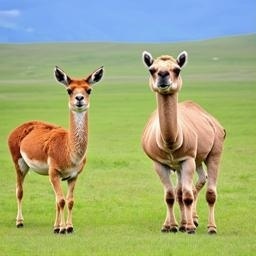}
			&\includegraphics[width=1.65cm]{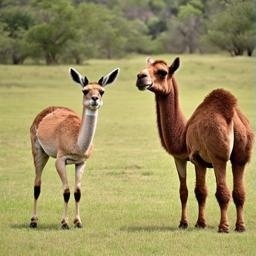}
			&\includegraphics[width=1.65cm]{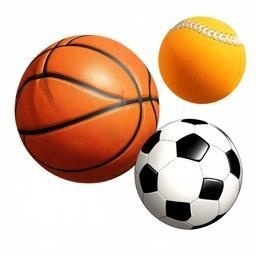}
			&\includegraphics[width=1.65cm]{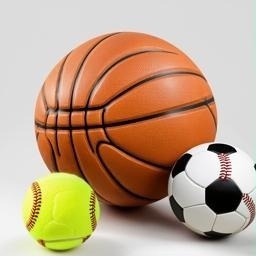}
			&\includegraphics[width=1.65cm]{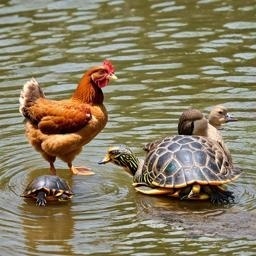}
			&\includegraphics[width=1.65cm]{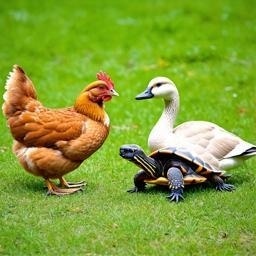}
			\\

			&\includegraphics[width=1.65cm]{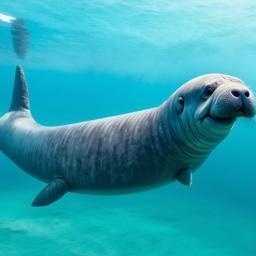}
			&\includegraphics[width=1.65cm]{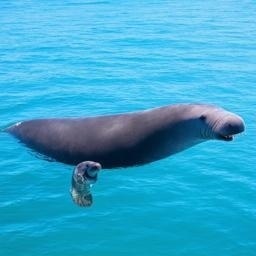}
			&\includegraphics[width=1.65cm]{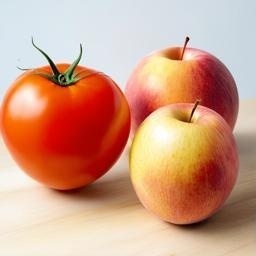}
			&\includegraphics[width=1.65cm]{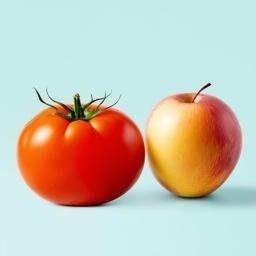}
			&\includegraphics[width=1.65cm]{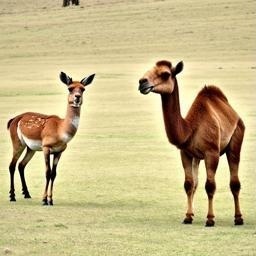}
			&\includegraphics[width=1.65cm]{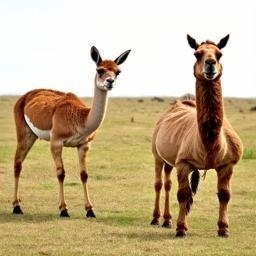}
			&\includegraphics[width=1.65cm]{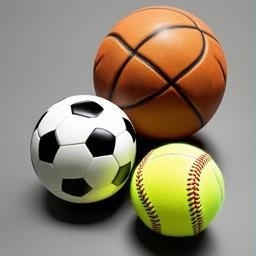}
			&\includegraphics[width=1.65cm]{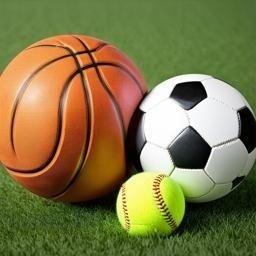}
			&\includegraphics[width=1.65cm]{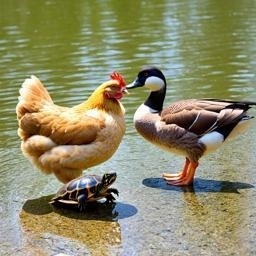}
			&\includegraphics[width=1.65cm]{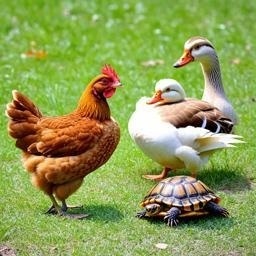}
			\\

                \\

			\multirow{2}{*}{\raisebox{-0.85cm}{\rotatebox[origin=c]{90}{\footnotesize{{A\&E}}}}}
			&\includegraphics[width=1.65cm]{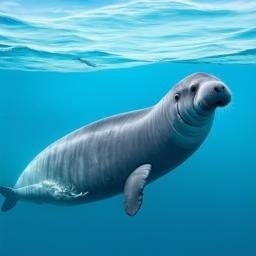}
			&\includegraphics[width=1.65cm]{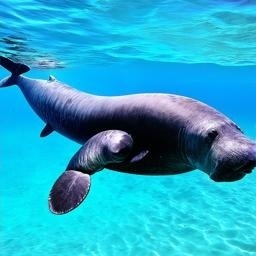}
			&\includegraphics[width=1.65cm]{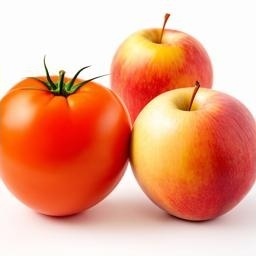}
			&\includegraphics[width=1.65cm]{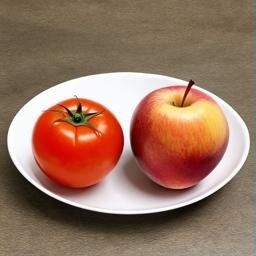}
			&\includegraphics[width=1.65cm]{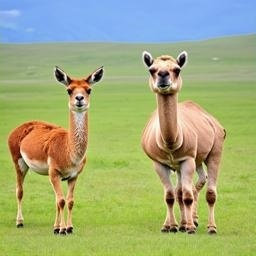}
			&\includegraphics[width=1.65cm]{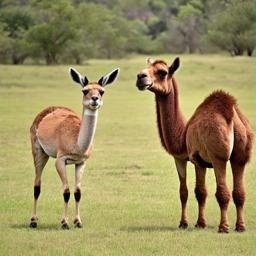}
			&\includegraphics[width=1.65cm]{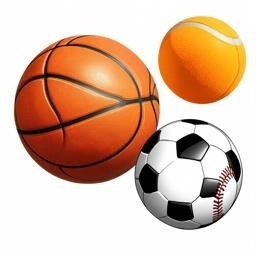}
			&\includegraphics[width=1.65cm]{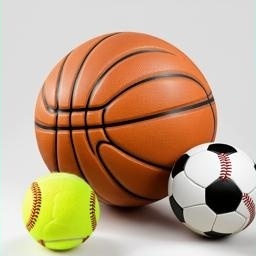}
			&\includegraphics[width=1.65cm]{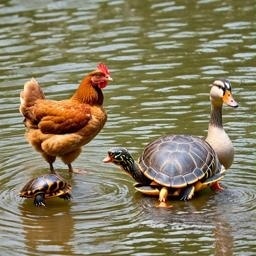}
			&\includegraphics[width=1.65cm]{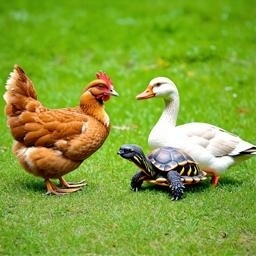}
			\\

			&\includegraphics[width=1.65cm]{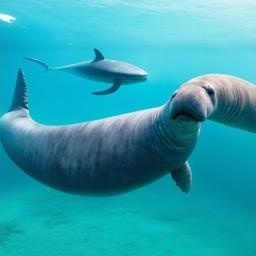}
			&\includegraphics[width=1.65cm]{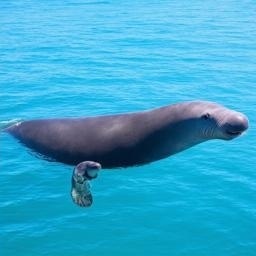}
			&\includegraphics[width=1.65cm]{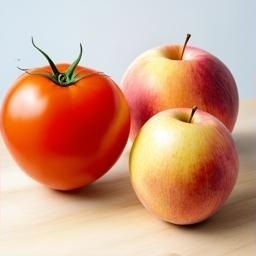}
			&\includegraphics[width=1.65cm]{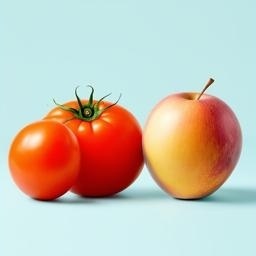}
			&\includegraphics[width=1.65cm]{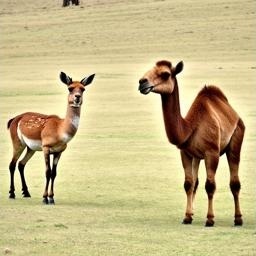}
			&\includegraphics[width=1.65cm]{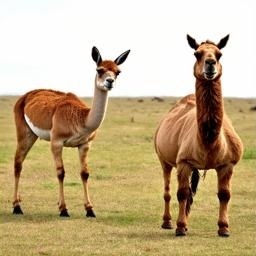}
			&\includegraphics[width=1.65cm]{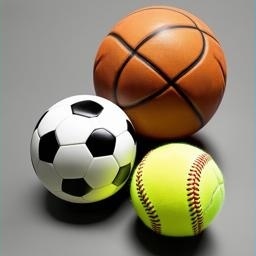}
			&\includegraphics[width=1.65cm]{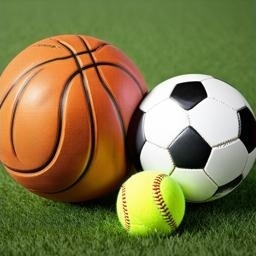}
			&\includegraphics[width=1.65cm]{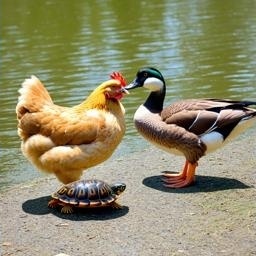}
			&\includegraphics[width=1.65cm]{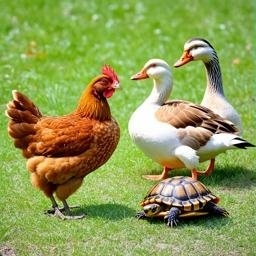}
			\\

                \\

			\multirow{2}{*}{\raisebox{-0.85cm}{\rotatebox[origin=c]{90}{\footnotesize{{EBAMA}}}}}
			&\includegraphics[width=1.65cm]{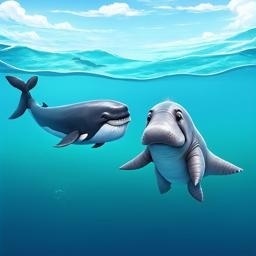}
			&\includegraphics[width=1.65cm]{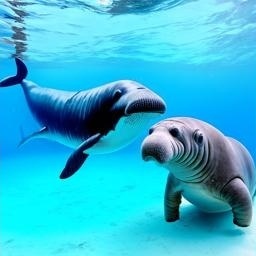}
			&\includegraphics[width=1.65cm]{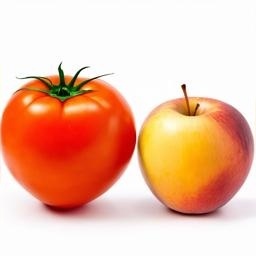}
			&\includegraphics[width=1.65cm]{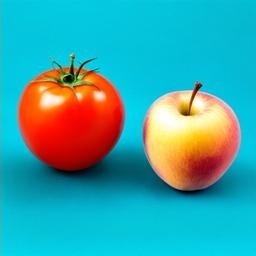}
			&\includegraphics[width=1.65cm]{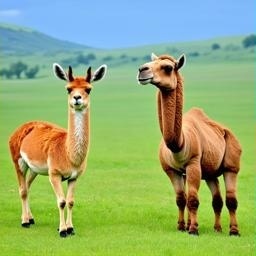}
			&\includegraphics[width=1.65cm]{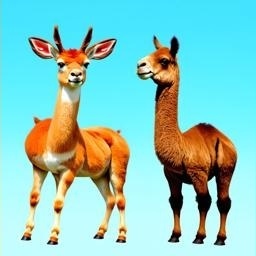}
			&\includegraphics[width=1.65cm]{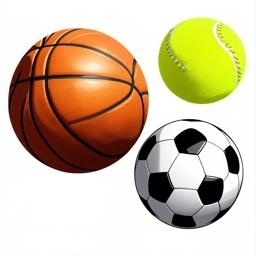}
			&\includegraphics[width=1.65cm]{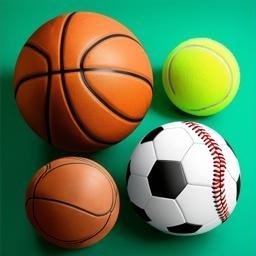}
			&\includegraphics[width=1.65cm]{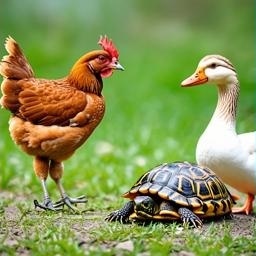}
			&\includegraphics[width=1.65cm]{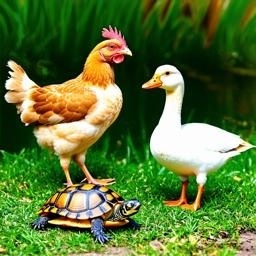}
			\\
		
			&\includegraphics[width=1.65cm]{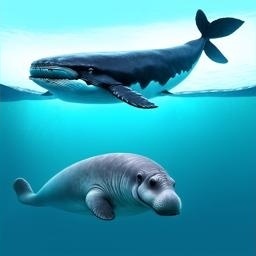}
			&\includegraphics[width=1.65cm]{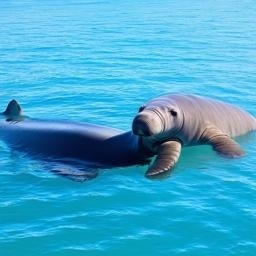}
			&\includegraphics[width=1.65cm]{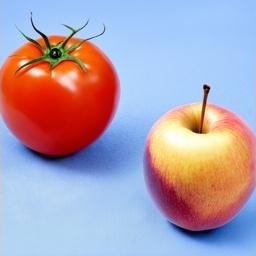}
			&\includegraphics[width=1.65cm]{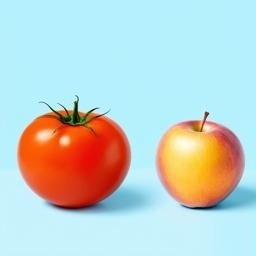}
			&\includegraphics[width=1.65cm]{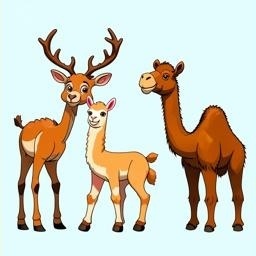}
			&\includegraphics[width=1.65cm]{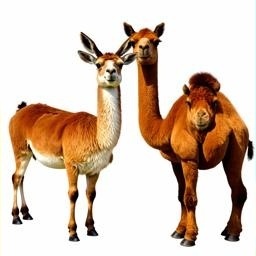}
			&\includegraphics[width=1.65cm]{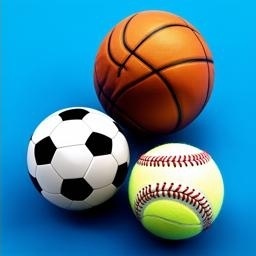}
			&\includegraphics[width=1.65cm]{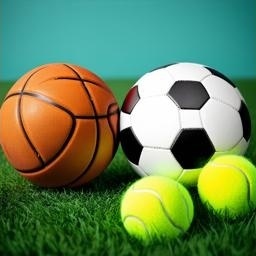}
			&\includegraphics[width=1.65cm]{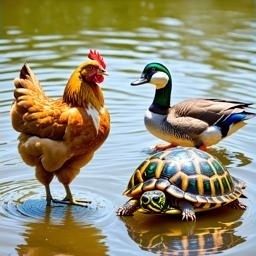}
			&\includegraphics[width=1.65cm]{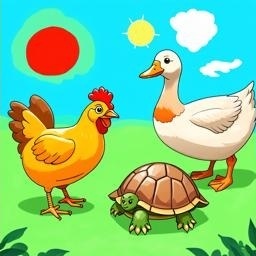}
			\\

                \\
                
			\multirow{2}{*}{\raisebox{-0.85cm}{\rotatebox[origin=c]{90}{\footnotesize{{CONFORM}}}}}
			&\includegraphics[width=1.65cm]{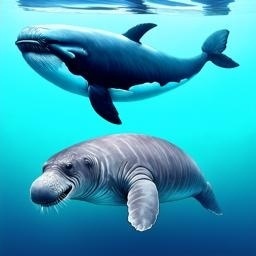}
			&\includegraphics[width=1.65cm]{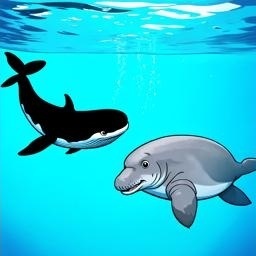}
			&\includegraphics[width=1.65cm]{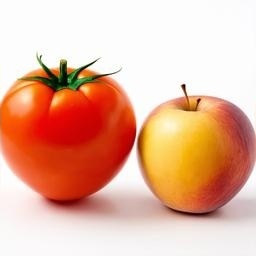}
			&\includegraphics[width=1.65cm]{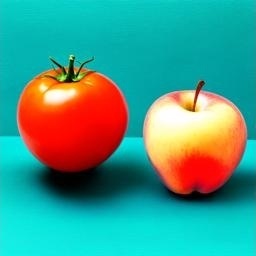}
			&\includegraphics[width=1.65cm]{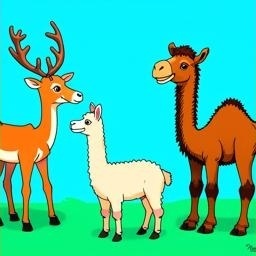}
			&\includegraphics[width=1.65cm]{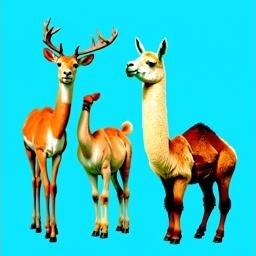}
			&\includegraphics[width=1.65cm]{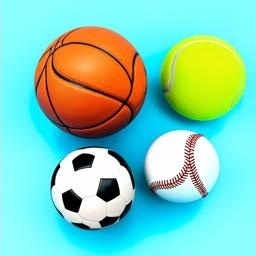}
			&\includegraphics[width=1.65cm]{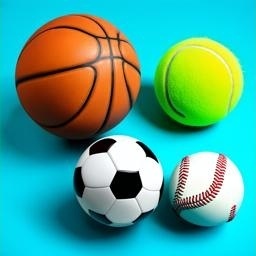}
			&\includegraphics[width=1.65cm]{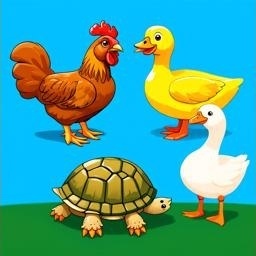}
			&\includegraphics[width=1.65cm]{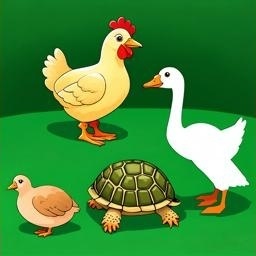}
			\\

			&\includegraphics[width=1.65cm]{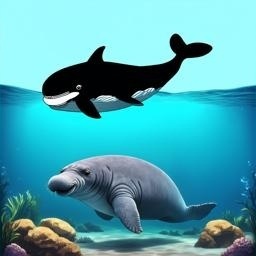}
			&\includegraphics[width=1.65cm]{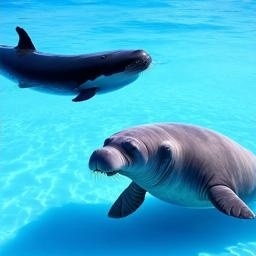}
			&\includegraphics[width=1.65cm]{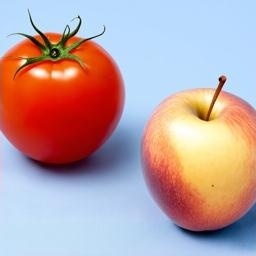}
			&\includegraphics[width=1.65cm]{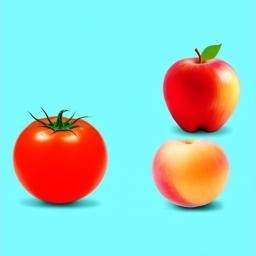}
			&\includegraphics[width=1.65cm]{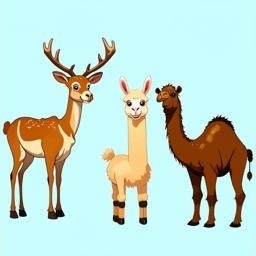}
			&\includegraphics[width=1.65cm]{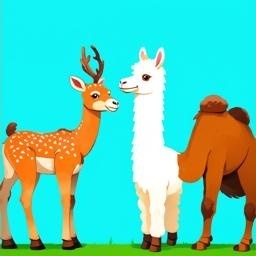}
			&\includegraphics[width=1.65cm]{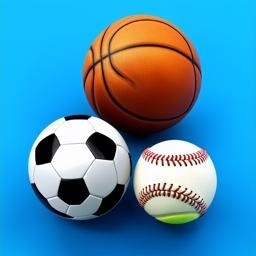}
			&\includegraphics[width=1.65cm]{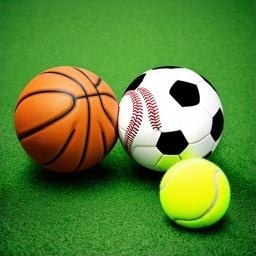}
			&\includegraphics[width=1.65cm]{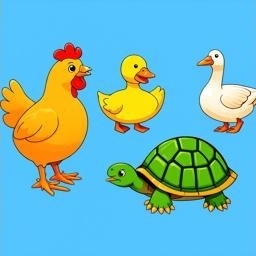}
			&\includegraphics[width=1.65cm]{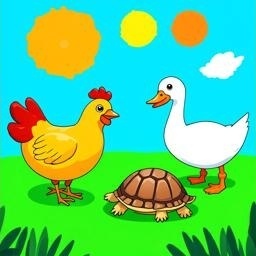}
			\\

                \\

			\multirow{2}{*}{\raisebox{-0.85cm}{\rotatebox[origin=c]{90}{\footnotesize{{Ours}}}}}
			&\includegraphics[width=1.65cm]{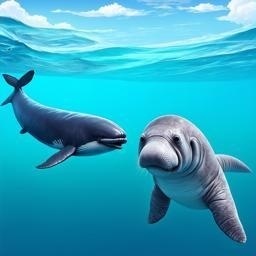}
			&\includegraphics[width=1.65cm]{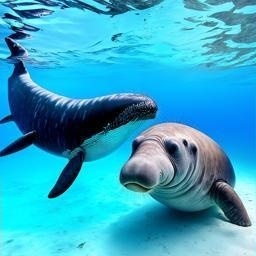}
			&\includegraphics[width=1.65cm]{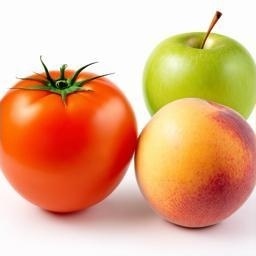}
			&\includegraphics[width=1.65cm]{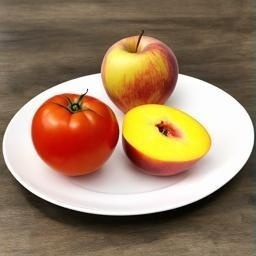}
			&\includegraphics[width=1.65cm]{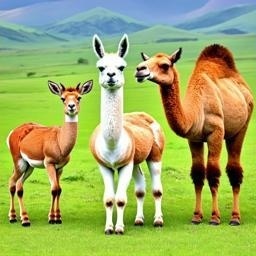}
			&\includegraphics[width=1.65cm]{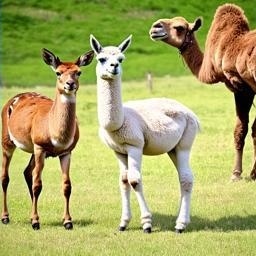}
			&\includegraphics[width=1.65cm]{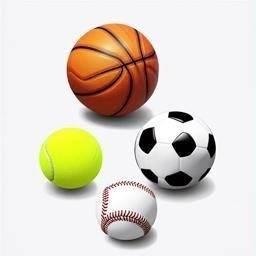}
			&\includegraphics[width=1.65cm]{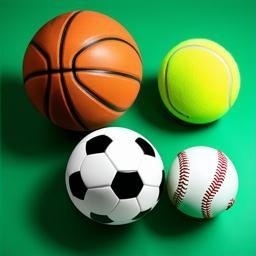}
			&\includegraphics[width=1.65cm]{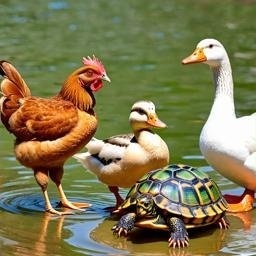}
			&\includegraphics[width=1.65cm]{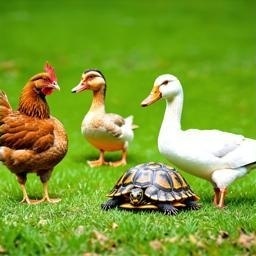}
			\\

			&\includegraphics[width=1.65cm]{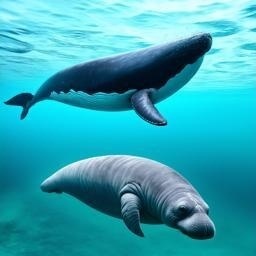}
			&\includegraphics[width=1.65cm]{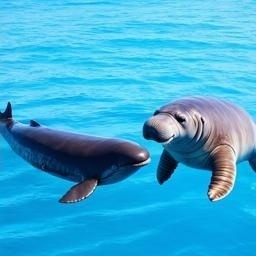}
			&\includegraphics[width=1.65cm]{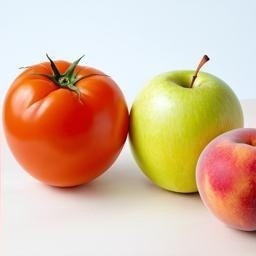}
			&\includegraphics[width=1.65cm]{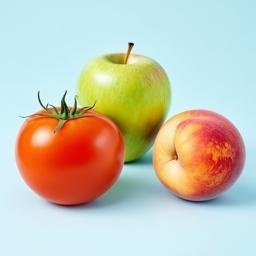}
			&\includegraphics[width=1.65cm]{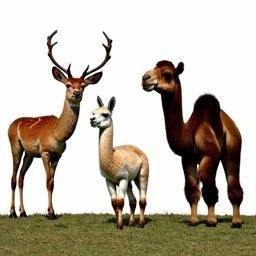}
			&\includegraphics[width=1.65cm]{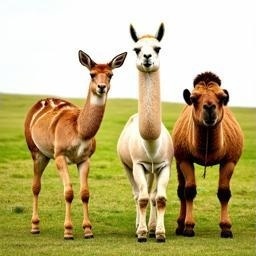}
			&\includegraphics[width=1.65cm]{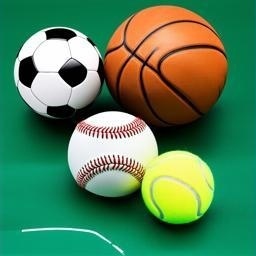}
			&\includegraphics[width=1.65cm]{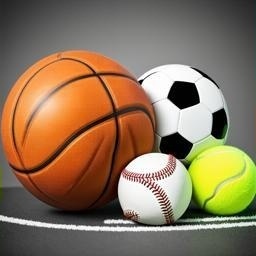}
			&\includegraphics[width=1.65cm]{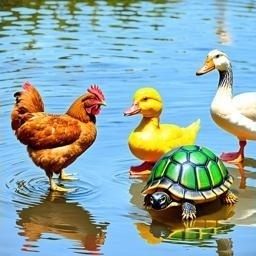}
			&\includegraphics[width=1.65cm]{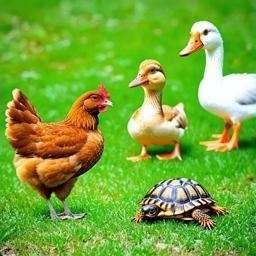}
			\\
            
		\end{tabular}
	\end{center}
	\caption{Qualitative comparison with A\&E~\cite{chefer2023attend}, EBAMA~\cite{zhang2024object}, CONFORM~\cite{meral2024conform} on prompts containing two, three, and four similar subjects. Our approach mitigates the subject neglect or mixing problems present in SD3~\cite{esser2024scaling} while maintaining the superior generation quality.}
	\label{fig:quail_comparsion_1}
\end{figure*}

\begin{figure*}[t]
	\begin{center}
		\setlength{\tabcolsep}{0.5pt}
		\begin{tabular}{m{0.3cm}<{\centering}m{1.68cm}<{\centering}m{1.68cm}<{\centering}m{1.68cm}<{\centering}m{1.68cm}<{\centering}m{1.68cm}<{\centering}m{1.68cm}<{\centering}m{1.68cm}<{\centering}m{1.68cm}<{\centering}m{1.68cm}<{\centering}m{1.68cm}<{\centering}}
			 & \multicolumn{2}{c}{\scriptsize{\textit{``a \underline{helicopter} and a \underline{warplane}"}}} & \multicolumn{2}{c}{\scriptsize{\makecell{\scriptsize{\textit{``a \underline{bear} and a \underline{wolf}}} \\ \scriptsize{\textit{and a \underline{fox}"}}}}} & \multicolumn{2}{c}{\scriptsize{\makecell{\scriptsize{\textit{``a \underline{carrot} and a \underline{potato}}} \\ \scriptsize{\textit{and a \underline{parsnip}"}}}}} & \multicolumn{2}{c}{\scriptsize{\makecell{\scriptsize{\textit{``an \underline{eagle} and a \underline{bird} and}} \\ \scriptsize{\textit{a \underline{butterfly} and a \underline{bee}"}}}}} & \multicolumn{2}{c}{\scriptsize{\makecell{\scriptsize{\textit{``a \underline{cherry} and a \underline{grape} and}} \\ \scriptsize{\textit{a \underline{plum} and a \underline{apricot}"}}}}}
			\\
                \noalign{\vskip 3pt}
   
			\multirow{2}{*}{\raisebox{-0.85cm}{\rotatebox[origin=c]{90}{\footnotesize{{SD3}}}}}
			&\includegraphics[width=1.65cm]{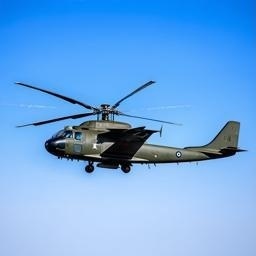}
			&\includegraphics[width=1.65cm]{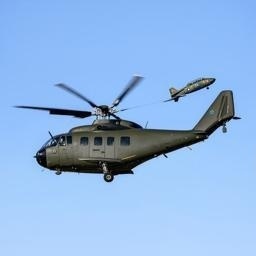}
			&\includegraphics[width=1.65cm]{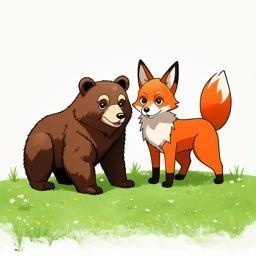}
			&\includegraphics[width=1.65cm]{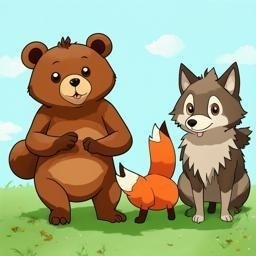}
			&\includegraphics[width=1.65cm]{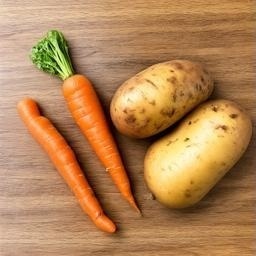}
			&\includegraphics[width=1.65cm]{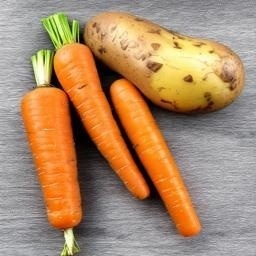}
			&\includegraphics[width=1.65cm]{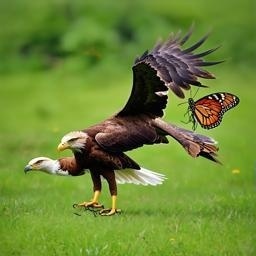}
			&\includegraphics[width=1.65cm]{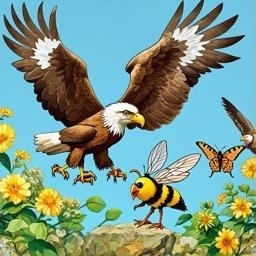}
			&\includegraphics[width=1.65cm]{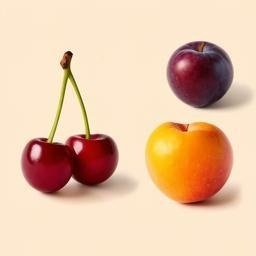}
			&\includegraphics[width=1.65cm]{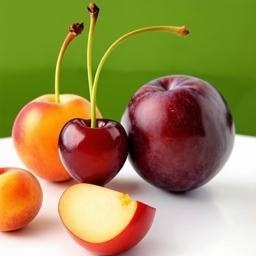}
			\\

			&\includegraphics[width=1.65cm]{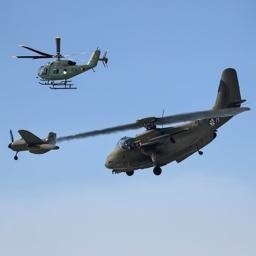}
			&\includegraphics[width=1.65cm]{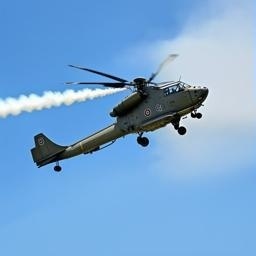}
			&\includegraphics[width=1.65cm]{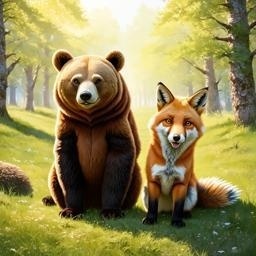}
			&\includegraphics[width=1.65cm]{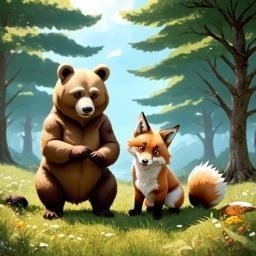}
			&\includegraphics[width=1.65cm]{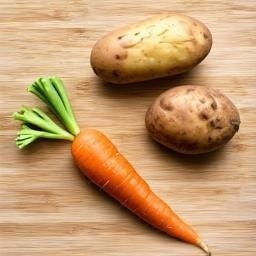}
			&\includegraphics[width=1.65cm]{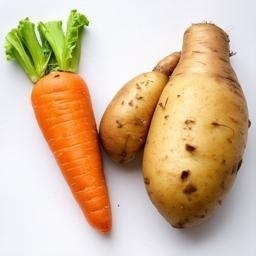}
			&\includegraphics[width=1.65cm]{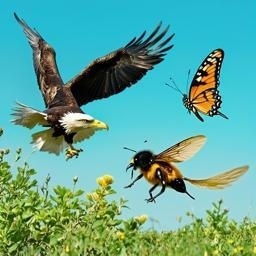}
			&\includegraphics[width=1.65cm]{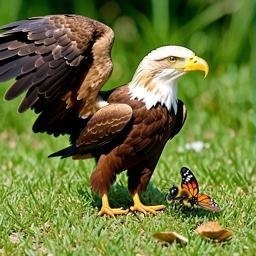}
			&\includegraphics[width=1.65cm]{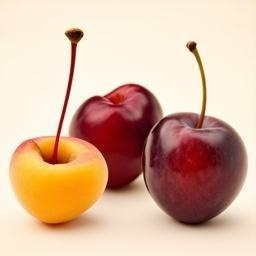}
			&\includegraphics[width=1.65cm]{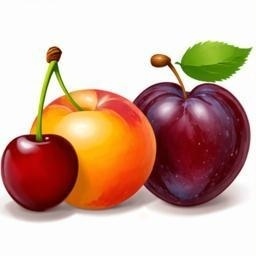}
			\\

                \\

			\multirow{2}{*}{\raisebox{-0.85cm}{\rotatebox[origin=c]{90}{\footnotesize{{A\&E}}}}}
			&\includegraphics[width=1.65cm]{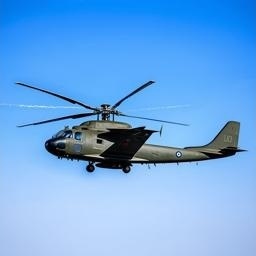}
			&\includegraphics[width=1.65cm]{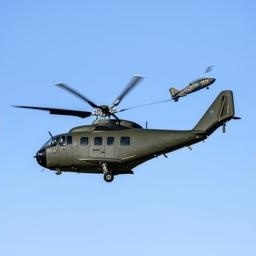}
			&\includegraphics[width=1.65cm]{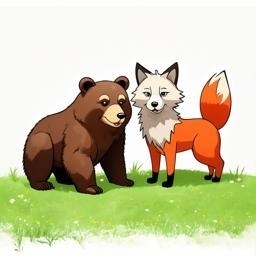}
			&\includegraphics[width=1.65cm]{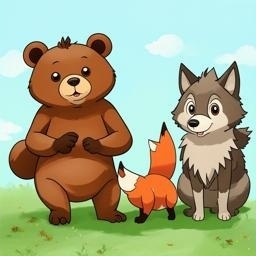}
			&\includegraphics[width=1.65cm]{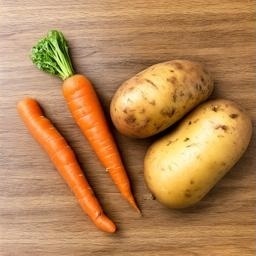}
			&\includegraphics[width=1.65cm]{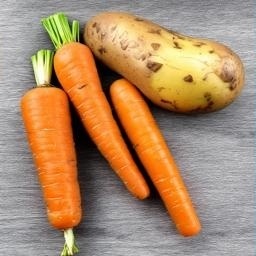}
			&\includegraphics[width=1.65cm]{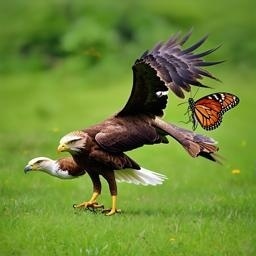}
			&\includegraphics[width=1.65cm]{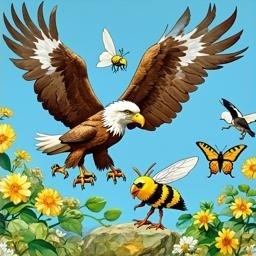}
			&\includegraphics[width=1.65cm]{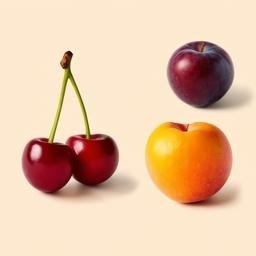}
			&\includegraphics[width=1.65cm]{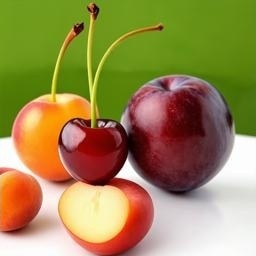}
			\\

			&\includegraphics[width=1.65cm]{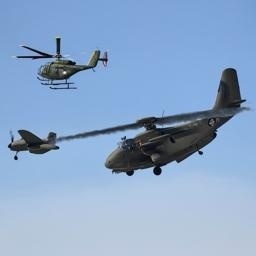}
			&\includegraphics[width=1.65cm]{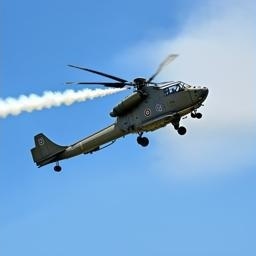}
			&\includegraphics[width=1.65cm]{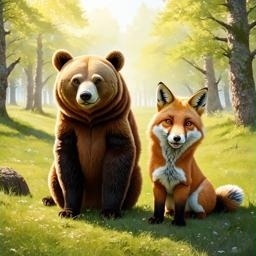}
			&\includegraphics[width=1.65cm]{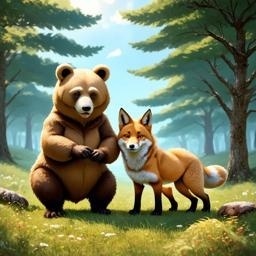}
			&\includegraphics[width=1.65cm]{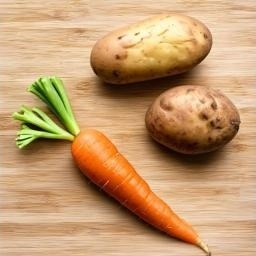}
			&\includegraphics[width=1.65cm]{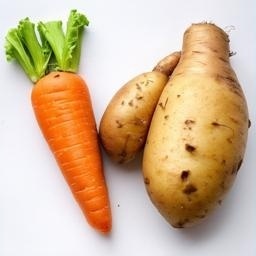}
			&\includegraphics[width=1.65cm]{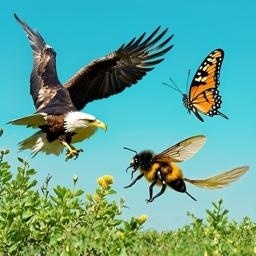}
			&\includegraphics[width=1.65cm]{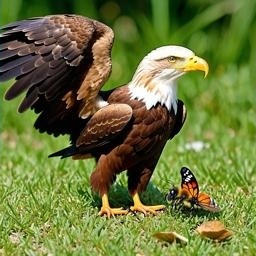}
			&\includegraphics[width=1.65cm]{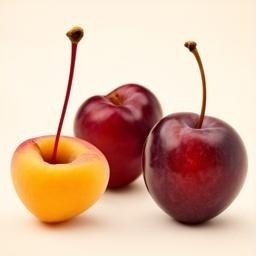}
			&\includegraphics[width=1.65cm]{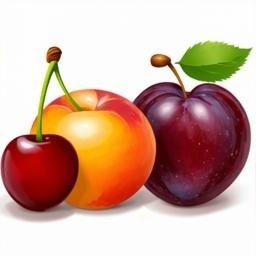}
			\\

                \\

			\multirow{2}{*}{\raisebox{-0.85cm}{\rotatebox[origin=c]{90}{\footnotesize{{EBAMA}}}}}
			&\includegraphics[width=1.65cm]{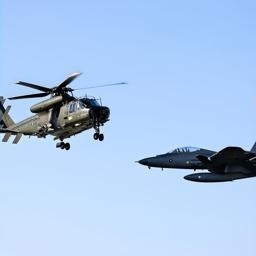}
			&\includegraphics[width=1.65cm]{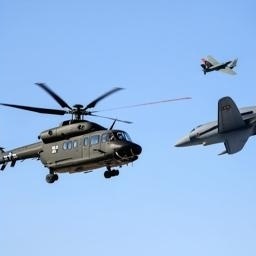}
			&\includegraphics[width=1.65cm]{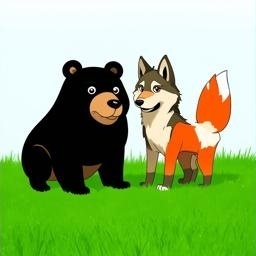}
			&\includegraphics[width=1.65cm]{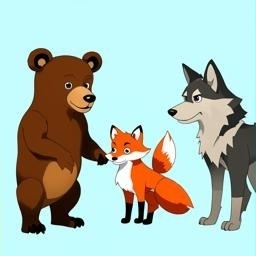}
			&\includegraphics[width=1.65cm]{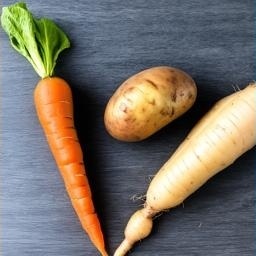}
			&\includegraphics[width=1.65cm]{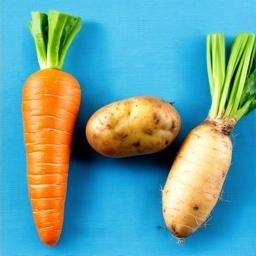}
			&\includegraphics[width=1.65cm]{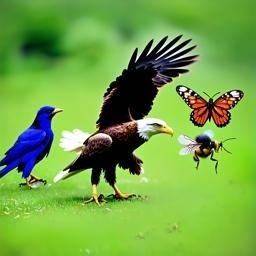}
			&\includegraphics[width=1.65cm]{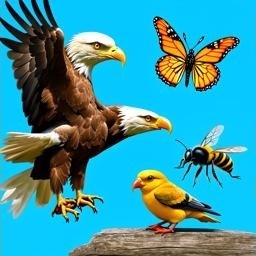}
			&\includegraphics[width=1.65cm]{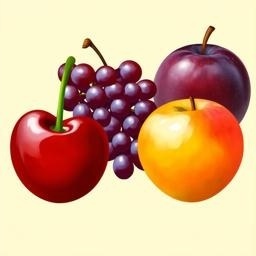}
			&\includegraphics[width=1.65cm]{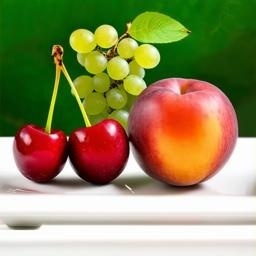}
			\\
		
			&\includegraphics[width=1.65cm]{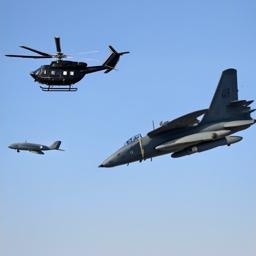}
			&\includegraphics[width=1.65cm]{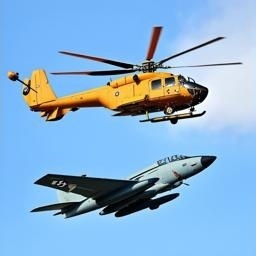}
			&\includegraphics[width=1.65cm]{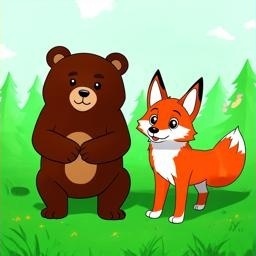}
			&\includegraphics[width=1.65cm]{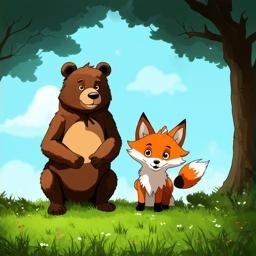}
			&\includegraphics[width=1.65cm]{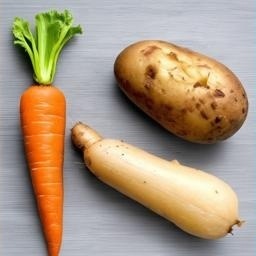}
			&\includegraphics[width=1.65cm]{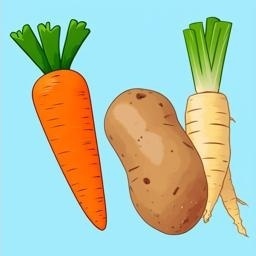}
			&\includegraphics[width=1.65cm]{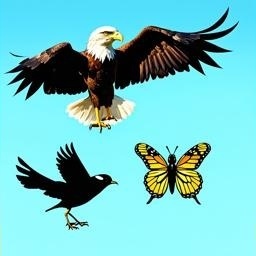}
			&\includegraphics[width=1.65cm]{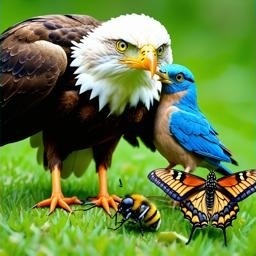}
			&\includegraphics[width=1.65cm]{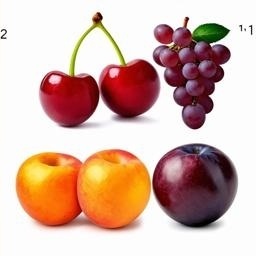}
			&\includegraphics[width=1.65cm]{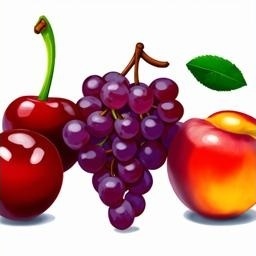}
			\\

                \\
                
			\multirow{2}{*}{\raisebox{-0.85cm}{\rotatebox[origin=c]{90}{\footnotesize{{CONFORM}}}}}
			&\includegraphics[width=1.65cm]{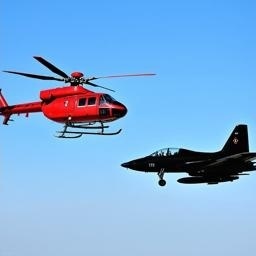}
			&\includegraphics[width=1.65cm]{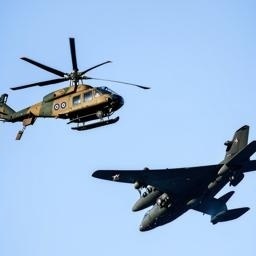}
			&\includegraphics[width=1.65cm]{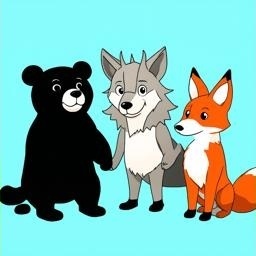}
			&\includegraphics[width=1.65cm]{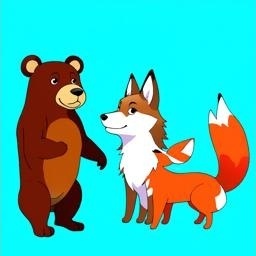}
			&\includegraphics[width=1.65cm]{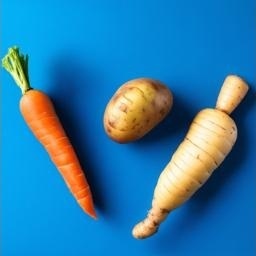}
			&\includegraphics[width=1.65cm]{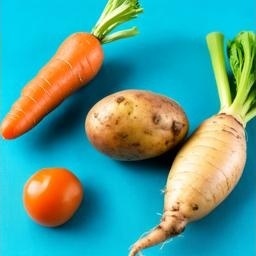}
			&\includegraphics[width=1.65cm]{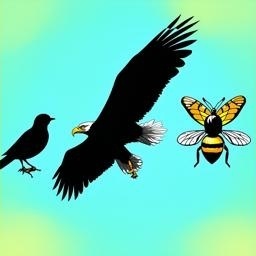}
			&\includegraphics[width=1.65cm]{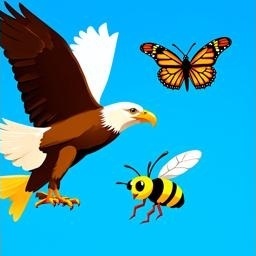}
			&\includegraphics[width=1.65cm]{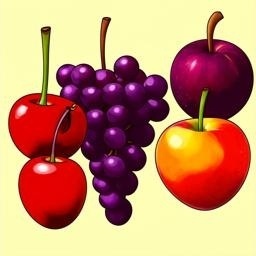}
			&\includegraphics[width=1.65cm]{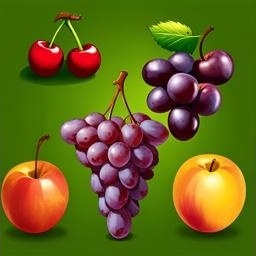}
			\\

			&\includegraphics[width=1.65cm]{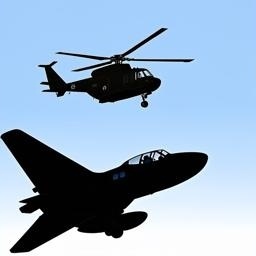}
			&\includegraphics[width=1.65cm]{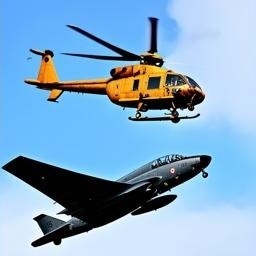}
			&\includegraphics[width=1.65cm]{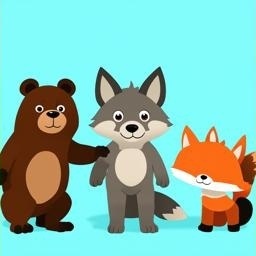}
			&\includegraphics[width=1.65cm]{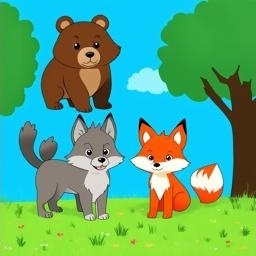}
			&\includegraphics[width=1.65cm]{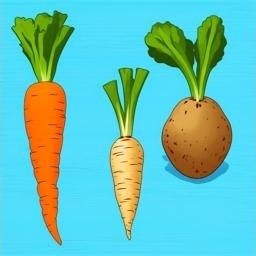}
			&\includegraphics[width=1.65cm]{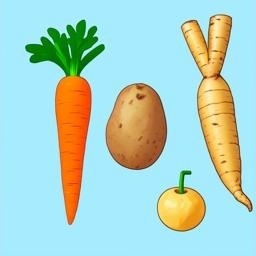}
			&\includegraphics[width=1.65cm]{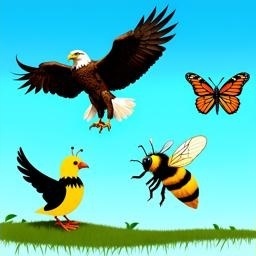}
			&\includegraphics[width=1.65cm]{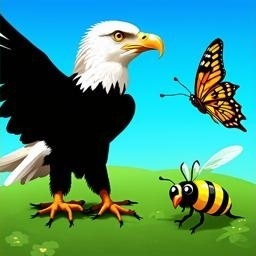}
			&\includegraphics[width=1.65cm]{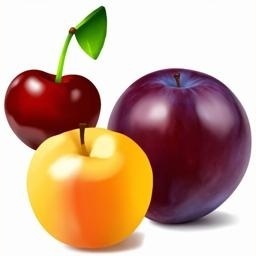}
			&\includegraphics[width=1.65cm]{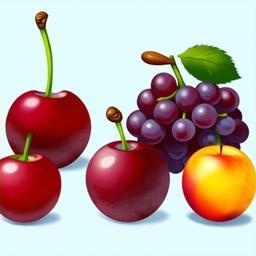}
			\\

                \\

			\multirow{2}{*}{\raisebox{-0.85cm}{\rotatebox[origin=c]{90}{\footnotesize{{Ours}}}}}
			&\includegraphics[width=1.65cm]{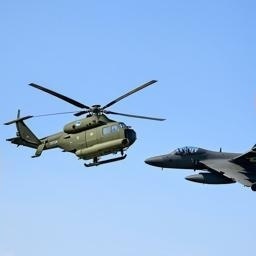}
			&\includegraphics[width=1.65cm]{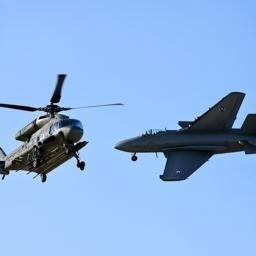}
			&\includegraphics[width=1.65cm]{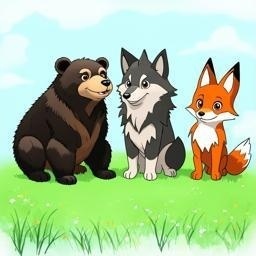}
			&\includegraphics[width=1.65cm]{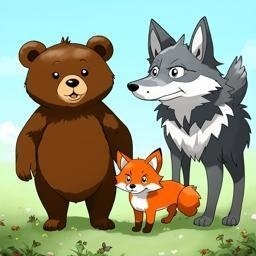}
			&\includegraphics[width=1.65cm]{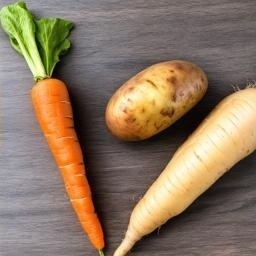}
			&\includegraphics[width=1.65cm]{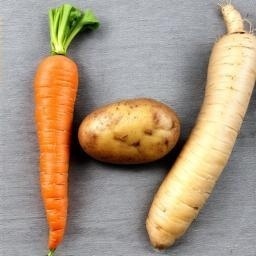}
			&\includegraphics[width=1.65cm]{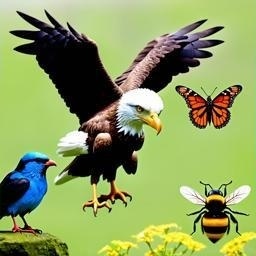}
			&\includegraphics[width=1.65cm]{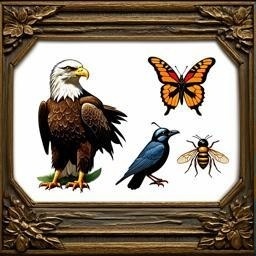}
			&\includegraphics[width=1.65cm]{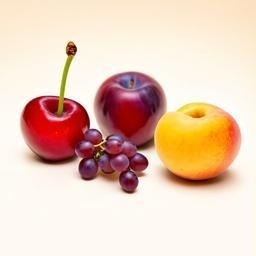}
			&\includegraphics[width=1.65cm]{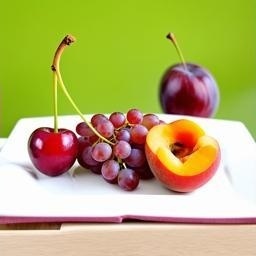}
			\\

			&\includegraphics[width=1.65cm]{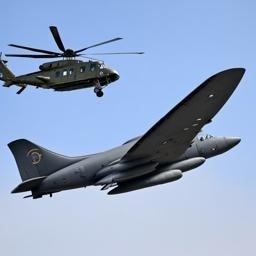}
			&\includegraphics[width=1.65cm]{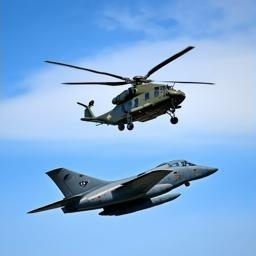}
			&\includegraphics[width=1.65cm]{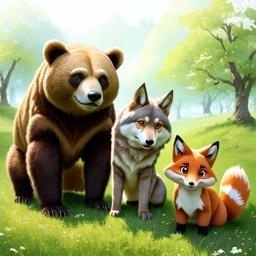}
			&\includegraphics[width=1.65cm]{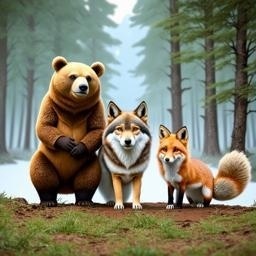}
			&\includegraphics[width=1.65cm]{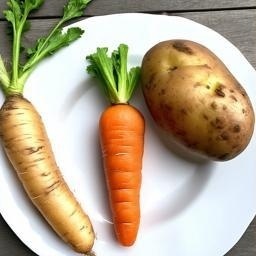}
			&\includegraphics[width=1.65cm]{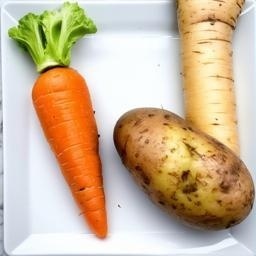}
			&\includegraphics[width=1.65cm]{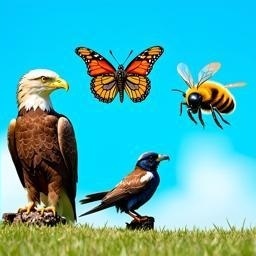}
			&\includegraphics[width=1.65cm]{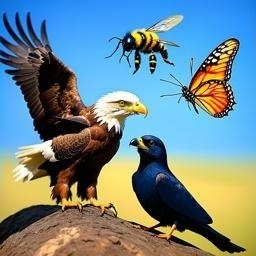}
			&\includegraphics[width=1.65cm]{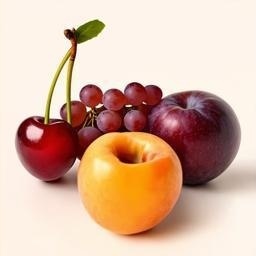}
			&\includegraphics[width=1.65cm]{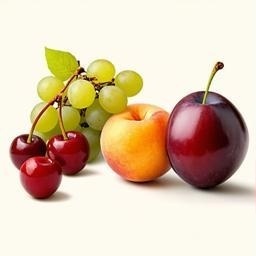}
			\\
            
		\end{tabular}
	\end{center}
	\caption{Qualitative comparison with A\&E~\cite{chefer2023attend}, EBAMA~\cite{zhang2024object}, CONFORM~\cite{meral2024conform} on prompts containing two, three, and four similar subjects. Our approach mitigates the subject neglect or mixing problems present in SD3~\cite{esser2024scaling} while maintaining the superior generation quality.}
	\label{fig:quail_comparsion_2}
\end{figure*}

\end{document}